\pdfoutput=1

\documentclass[twocolumn]{article}
\usepackage{acl}

\usepackage{authblk}
\usepackage[T1]{fontenc}
\usepackage[utf8]{inputenc}
\usepackage{times}
\usepackage{latexsym}
\usepackage{CJKutf8}
\usepackage{inconsolata}
\usepackage{microtype}

\usepackage{amsmath}
\usepackage{amssymb}

\usepackage{graphicx}
\usepackage{xcolor}
\usepackage{colortbl}
\definecolor{c2}{rgb}{1,0,0}
\usepackage{subcaption}
\usepackage{booktabs}
\usepackage{array}
\usepackage{multirow}
\usepackage{adjustbox}
\usepackage{arydshln}
\usepackage{tabularx}
\usepackage{placeins}

\usepackage[ruled,vlined]{algorithm2e}

\usepackage[shortlabels]{enumitem}
\usepackage{stfloats}

\title{MultiTEND: A Multilingual Benchmark for Natural Language to NoSQL Query Translation}

\author{Zhiqian Qin$^{1}$, Yuanfeng Song$^{2}$\thanks{$^{\ast}$
Yuanfeng Song is the corresponding author.}, Jinwei Lu$^1$, Yuanwei Song$^3$, Shuaimin Li$^1$, Chen Jason Zhang$^1$ \\
$^1$The Hong Kong Polytechnic University, Hong Kong SAR, China \\
$^2$AI Group, WeBank Co., Ltd, Shenzhen, China \\
$^3$Huawei Technologies Ltd. \\
\\}

\begin{document}
\maketitle
\begin{abstract}
Natural language interfaces for NoSQL databases are increasingly vital in the big data era, enabling users to interact with complex, unstructured data without deep technical expertise. However, most recent advancements focus on English, leaving a gap for multilingual support. This paper introduces \textbf{MultiTEND}, the first and largest multilingual benchmark for natural language to NoSQL query generation, covering six languages: English, German, French, Russian, Japanese and Mandarin Chinese.
{Using MultiTEND, we analyze challenges in translating natural language to NoSQL queries across diverse linguistic structures, including lexical and syntactic differences. Experiments show that performance accuracy in both English and non-English settings remains relatively low, with a 4\%-6\% gap across scenarios like fine-tuned SLM, zero-shot LLM, and RAG for LLM.
To address the aforementioned challenges, we introduce \textbf{MultiLink}, a novel framework that bridges the multilingual input to NoSQL query generation gap through a Parallel Linking Process. It breaks down the task into multiple steps, integrating parallel multilingual processing, Chain-of-Thought (CoT) reasoning, and Retrieval-Augmented Generation (RAG) to tackle lexical and structural challenges inherent in multilingual NoSQL generation. MultiLink shows enhancements in all metrics for every language against the top baseline, boosting execution accuracy by about 15\% for English and averaging a 10\% improvement for non-English languages.}

\end{abstract}

\section{Introduction}

In the age of big data, NoSQL databases have become indispensable tools for managing vast amounts of unstructured and semi-structured data~\cite{6106531}. Unlike traditional relational databases, NoSQL databases offer more flexibility in schema design and can handle a wide variety of data types, making them particularly suitable for modern applications such as social media~\cite{7096207}, e-commerce~\cite{nalla2022sql}, and real-time analytics~\cite{7529531}. Nonetheless, the intricacy and heterogeneity of NoSQL query languages present a formidable challenge, especially for users who may not have advanced technical skills.

To address this challenge, the development of natural language interfaces (NLIs) for NoSQL databases has gained increasing attention. These interfaces are designed to allow users to interact with NoSQL databases in natural language, thus simplifying access to complex data and lowering the technical barriers. By translating Natural Language Queries (NLQs) into executable NoSQL queries (i.e., Text-to-NoSQL \cite{lu2024natural}), these systems can significantly enhance user productivity and data accessibility.
However, existing natural language to NoSQL query generation systems and benchmarks have predominantly focused on the English language. This limitation severely restricts the usability of these systems for non-English speakers, who represent a significant portion of the global population. 

\begin{figure*}[th!]
    \centering  
    \includegraphics[width=1.0\textwidth]{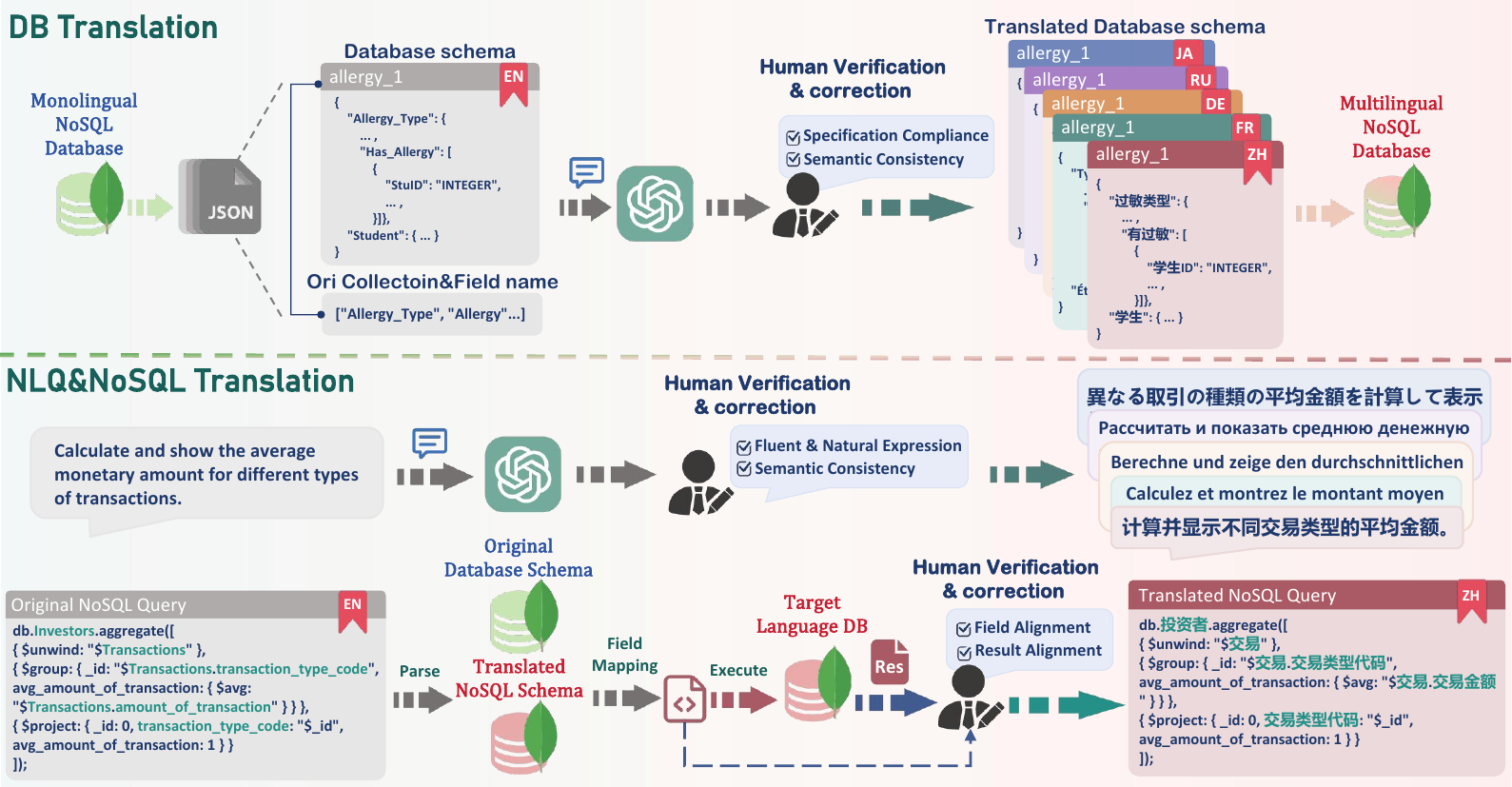}
    \vspace{-10pt}
    \caption{We developed a semi-automated pipeline to extend the monolingual dataset into a multilingual version through three steps: (1) Translation of Database Fields, where English-exclusive fields were translated using LLM-powered tools and manually verified; (2) Translation of NLQs, where NLQs were translated with few-shot prompting for semantic consistency and manually corrected; and (3) Translation of NoSQL Queries, where queries were programmatically parsed, updated with multilingual representations, and verified based on execution results. Each step combined machine-generated methods with rigorous manual verification.}

    \label{fig:Dataset_pipeline}
    \vspace{-15pt}
\end{figure*}

To address above-mentioned issue, we introduce \textbf{MultiTEND}, the first multilingual benchmark for natural language to NoSQL query generation, covering six diverse languages: English, German, French, Russian, Japanese, and Mandarin Chinese (Sec.~\ref{sec Dataset: Overview}).
MultiTEND not only expands the scope of natural language to NoSQL query generation to a multilingual context but also imposes additional challenges to the Text-to-NoSQL tasks. Based on the findings from our experiments (Sec.~\ref{Sec Statistics: Findings}), We categorize the challenges in MultiTEND into Structural Challenge and Lexical Challenge. In particular, the Structural Challenge refers to difficulties models face in multilingual intention mapping tasks due to syntactic differences across languages, hindering accurate mapping to NoSQL operators. Additionally, the Lexical Challenge represents the schema linking difficulties models face in multilingual settings due to lexical differences (e.g., Japanese hiragana and katakana, Russian Cyrillic characters, and morphological variations in German and French) and the complexity of NoSQL structures (e.g., nested documents and array processing). 

To tackle these challenges, we propose \textbf{MultiLink}, a novel framework designed to bridge the gap from multilingual input to NoSQL query generation. Specifically, MultiTEND extracts accurate operator sketches and relevant fields from multilingual NLQs through a parallel linking process, enabling the model to effectively generate high-quality NoSQL queries even in multilingual contexts. 
Through the incorporation of three novel and specifically designed components, namely Intention-aware Multilingual Data Augmentation, Parallel Multilingual Sketch-Schema Prediction, and Retrieval-Augmented Chain-of-Thought Query Prediction, MultiLink effectively generates high-quality NoSQL queries tailored for multilingual scenarios.

In summary, our contributions are as follows:
\begin{itemize}[noitemsep, topsep=0pt, partopsep=0pt, parsep=0pt]
    \item We present MultiTEND, the largest multilingual benchmark for natural language to NoSQL query generation, which includes detailed construction methods and will be released to promote further research in this area.
    \item We conduct detailed analysis on the MultiTEND dataset, identifying the lexical and structural challenges in multilingual NoSQL generation, which arise from lexical and syntactic differences across languages as well as the inherent structural complexity of NoSQL queries.
    \item We introduce MultiLink, a novel framework designed for multilingual NoSQL query generation. By addressing both lexical and structural challenges through three innovative components, MultiLink achieves significantly better performance compared to other baselines in multilingual noSQL generation scenarios.
    \item We conduct extensive experiments on MultiTEND, demonstrating its challenging nature and the effectiveness of our proposed model in addressing these challenges.
\end{itemize}

The rest of this paper is structured as follows: Section 2 reviews related work in the field. Section 3 describes the details of the MultiTEND dataset. Section 4 outlines the architecture and training of the MultiLink model. Section 5 presents the experimental setup and results. Finally, Section 6 concludes the paper and discusses future directions.

\section{The MultiTEND Dataset}
\subsection{Overview}
\label{sec Dataset: Overview}
To address the limitation of existing datasets in the Text-to-NoSQL domain being solely constructed in English \cite{lu2024natural}, we propose the first and the largest multilingual benchmark MultiTEND in this field, covering six languages: English, German, French, Russian, Japanese, and Mandarin Chinese. In this section, we'll introduce the dataset construction pipeline (Sec \ref{sec Dataset: Construction Pipeline}) and the manual correction processing (Sec \ref{sec Dataset: Manual Correction}).

\begin{table*}[th!]
\centering
\scalebox{0.8}{
\begin{tabular}{l|lccccc}
\toprule
\rowcolor{gray!20} 
\textbf{Dataset} & \textbf{Input} & \textbf{Output} & \textbf{Query Source} & \textbf{Query Size} & \textbf{Languages} \\
\midrule
Spider \cite{yu-etal-2018-spider} & NLQ & SQL Query & Human-Labeled & \cellcolor{red!45!yellow!10}5693 & English \\
nvBench \cite{10.1145/3448016.3457261} & NLQ & Data Vis Query & Rule-based Synthesised & \cellcolor{red!45!yellow!21}7247 & English \\
OverpassQL \cite{staniek-etal-2024-text} & NLQ & Spatial Query & Crowdsourcing collected & 3890 & English \\
TEND \cite{lu2024natural} & NLQ & NoSQL Query & Machine generated \checkmark & 3308 & English \\
\cmidrule{1-6}
MultiTEND & NLQ & NoSQL Query & Machine generated \& Human Check \checkmark & \cellcolor{red!45!yellow!35}19848 & Multiple(6) \checkmark \\
\bottomrule
\end{tabular}
}
\caption{Comparison of MultiTEND with Other Existing Benchmarks in Natural Language Interface fields.}
\vspace{-10pt}
\label{Table:DatasetTable}
\end{table*}

\subsection{Dataset Construction Pipeline}
\label{sec Dataset: Construction Pipeline}
We segment the dataset's translation content into DB fields, NLQs, and NoSQL queries, employing a combination of prompt engineering \cite{sahoo2024systematic} and manual corrections to construct the dataset.
\paragraph{Translation of DB Fields}
We interpret the task of translating database fields as obtaining relationship maps from English to five different languages for the database fields. We encapsulate instructions and contextual information conducive to accurate translation, such as the database schema, the fields to be translated, and the required output format into prompts (As shown in Appendix \ref{appendix:prompt in dataset}) and utilize a large language model (LLM) to complete the translation process. The translation results undergo detailed human inspection and correction (as shown in Section \ref{sec Dataset: Manual Correction}), ultimately yielding five maps from English to each target language for every database. These well-checked maps are used for the translation of the databases, resulting in a total of 924 databases covering six languages, derived from the original 154 English-language databases (Figure \ref{fig:Dataset_pipeline}).
\paragraph{Translation of NLQs}
We have established the following requirements for the translation of the NLQs: (i) Semantic alignment; (ii) Preservation of specific referenced values; (iii) Fluency in language expression. Among these, the requirement to preserve specific referenced values corresponds to multilingual database fields, where the actual database values remain consistent with the original English database to ensure data consistency. To achieve efficient and high-quality NLQ translation, we have designed a step-by-step, query-intent-based, structured Prompt with contextual examples for multilingual NLQ translation (Appendix \ref{appendix:prompt in dataset}). By encouraging the model to think step-by-step, we ensure the fluency and accuracy of the translation. Finally, we perform manual verification and correction of the generated NLQ (Sec \ref{sec Dataset: Manual Correction}) to ensure that the translated NLQ meets the specified requirements.

\paragraph{Translation of NoSQL Queries}
As mentioned earlier, for each database, we have already obtained five mapping tables from English to target language for the field names (Figure \ref{fig:Dataset_pipeline}) and conducted manual reviews and corrections. Based on these well-reviewed mapping tables, we mapped the fields referenced in each NoSQL query from English to five different languages. For each translated NoSQL query, we first filter out incorrect queries by executing them and checking the execution results, then apply manual corrections to strictly ensure the accuracy of the translated NoSQL queries.
\definecolor{row_color}{RGB}{238, 194, 205}
\definecolor{low_score}{rgb}{0.3, 0.5, 0.8}
\definecolor{high_score}{RGB}{183,034,048}

\subsection{Manual Correction}
\label{sec Dataset: Manual Correction}
\paragraph{Typical Errors Analysis}
\begin{CJK*}{UTF8}{gbsn}
As mentioned in Sec \ref{sec Dataset: Construction Pipeline}, we completed the translation of db fields, NLQs, and NoSQL queries empowered by LLM and conducted final manual inspections and corrections. We summarize some of the typical mistakes discovered during the inspection of translation process as follows.
For the translation part of DB fields, we identified two typical error categories: Polysemy and Abbreviation. \textbf{Polysemy}: Some fields can have different meanings depending on the database scenario, which is one reason for the inappropriate translation of certain database fields. For example, the term `Movements' in the `Aircraft\textunderscore{}Movements' field could refer to `motion,' `movement,' or more specifically, `take-offs and landings.' By analyzing the data type and specific values of this field, it becomes evident that `take-offs and landings' is the most suitable meaning within the context of aircraft operations. \textbf{Abbreviation}: Translating abbreviations in database fields, taking into account the database context, is inherently a challenging task. Such errors constituted a larger proportion of the issues we detected in the translation of db fields. For example, `fname' might be incorrectly translated as `f姓名' whereas it should be translated as `名' when considering its neighbor `lname'; similarly, `f\textunderscore{}id' could mean `flight ID' or `file ID,' depending on the theme of the collection. Similarly, `HS' from the `soccer\textunderscore{}2' database could stand for `High School,' `Home State,' or `Historical Score.' However, upon examining the neighboring fields and specific values within the collection, it turns out that `HS' actually means `Historical Score.'
In the translation part of NLQs, we found that most of the NLQs not meeting the requirements mentioned in Sec \ref{sec Dataset: Construction Pipeline} were due to insufficient fluency in the language, such as translating ``How many papers are `Atsushi Ohori' the author of?'' into ``有多少论文是`Atsushi Ohori'的作者?''. This result comes from directly translating `are' and `of' without considering the overall structure of the sentence.

\begin{table*}[th!]
\centering
{
\begin{tabular}{l|lccccccc}
\toprule
\rowcolor{row_color}
\textbf{Metric} & \textbf{Model} & \textbf{EN} & \textbf{ZH} & \textbf{FR} & \textbf{DE} & \textbf{JA} & \textbf{RU} & \textbf{AVG (5 langs)} \\
\midrule
\multirow{4}{*}{EM} 
 & Fine-tuned Llama & 17.05\% & 13.57\% & 16.53\% & 15.78\% & 16.40\% & 14.51\% & 15.36\% \\
 & Zero-shot LLM & \color{low_score}0.29\% & 0.61\% & 0.61\% & 0.54\% & 0.54\% & 0.29\% & \color{low_score}0.52\% \\
 & RAG for LLM & 16.09\% & 13.98\% & 15.62\% & 14.33\% & 12.02\% & 13.89\% & 13.97\% \\
 & SMART & \color{high_score}18.85\% & 13.94\% & 18.38\% & 18.30\% & 18.05\% & 15.89\% & \color{high_score}16.91\% \\
\cmidrule{1-9}
\multirow{4}{*}{EX} 
 & Fine-tuned Llama & 44.61\% & 36.86\% & 41.26\% & 41.44\% & 43.32\% & 38.23\% & 40.22\% \\
 & Zero-shot LLM & \color{low_score}36.58\% & 28.99\% & 33.86\% & 34.91\% & 30.63\% & 29.68\% & \color{low_score}31.61\% \\
 & RAG for LLM & \color{high_score}51.70\% & 47.02\% & 49.28\% & 48.59\% & 45.12\% & 45.99\% & \color{high_score}47.20\% \\
 & SMART & 48.86\% & 38.05\% & 44.69\% & 44.22\% & 43.30\% & 41.03\% & 42.26\% \\
\bottomrule
\end{tabular}
}

\caption{Comparison of Exact Match and Execution Accuracy results for each model across different languages on MultiTEND. Notice that \textbf{AVG} is the average value of the corresponding metric across the 5 \textbf{non-English} languages}
\label{tab:Performence Comparison of baselines on MultiTEND}

\end{table*}

\paragraph{Correction Criteria}
Based on the error cases observed during the aforementioned manual inspection process, we made several adjustments to the dataset aiming to ensure the following aspects: DB fields adhere to standard database design rules and feature precise translations of polysemous words and abbreviations, aligning them with the context of the database; NLQs maintain semantic consistency and are expressed fluently and naturally; and NoSQL query results are fully consistent with the original queries.
For example, we carefully examined some abbreviated fields in the original database to ensure that these abbreviations, which are difficult to understand without context, accurately convey their original meanings after translation (e.g., ``flno'' remains equivalent to ``flight number'' after translation). Additionally, we paid special attention to potential field duplication issues after translation, particularly for fields originally distinguished by case or singular/plural forms, as such differences might result in identical expressions in the target language. For example, the collection name ``continents'' and the field name ``Continent'' might both be translated as ``洲'' in Chinese; similarly, the collection name ``city'' and the field name ``City'' might both be translated as ``都市'' in Japanese.
\end{CJK*}

\section{Dataset Statistics and Analysis}
\label{Sec Statistics}

\subsection{Statistics of MultiTEND}
\label{Sec Statistics: MultiTEND}

After our multilingual extension of TEND~\cite{lu2024natural}, MultiTEND ultimately includes a total of 154 databases with different content, comprising 924 databases in total, and 101,789 (NLQ, NoSQL) pairs (including 20,351 distinct NoSQL queries). The count of 101,789 pairs is derived from the fact that each query corresponds to five NLQs, with each NLQ further represented in six language versions. Approximately 16.6\% of all NoSQL queries use the find method (which includes filter, projection, sort, and limit operations), while the remaining queries use the aggregate method (implemented through pipelines, which include but are not limited to project, group, match, sort, limit, lookup, and count stages) (Detailed statistics of MultiTEND see Apendix \ref{appendix: Dataset Details}).
Compared to other well-regarded datasets in different fields, such as Spider \cite{yu-etal-2018-spider}, NvBench \cite{luo2021synthesizing}, OverpassQL \cite{staniek-etal-2024-text}, and TEND \cite{lu2024natural}, MultiTEND stands out for its vast scale and comprehensive multilingual coverage (as shown in Table \ref{Table:DatasetTable}). In terms of scale, MultiTEND boasts 101,789 NLQs and 20,351 corresponding queries, far surpassing other datasets like Spider, which has 10,181 NLQs and 5,693 queries; NvBench, with 25,750 NLQs and 7,247 queries; and TEND, featuring 17,020 NLQs and 3,404 queries. Regarding multilingual support, unlike other datasets that primarily offer data in English, MultiTEND supports six distinct languages, greatly broadening its applicability and research value. Additionally, MultiTEND's semi-automated construction process, which combines machine-generated data with manual verification, provides significant advantages in terms of scalability and efficiency during its development.

\subsection{Analysis and Findings}
\label{Sec Statistics: Findings}
To clarify the challenges posed by multilingual Text-to-NoSQL tasks for existing models, we conducted a series of experiments and derived key findings from the analysis of the experimental results (See Appendix \ref{appendix: Analysis and Findings}).
Based on the findings, we categorize the challenges in MultiTEND into \textbf{Structural Challenge} and \textbf{Lexical Challenge}. (i) The \textbf{Structural Challenge} refers to the difficulties models face in performing intention mapping tasks in multilingual contexts, primarily due to significant syntactic differences across languages, which reduce the model's ability to understand and parse user intentions, making it harder to accurately map them to corresponding NoSQL operators. (ii) The \textbf{Lexical Challenge} refers to the difficulties models encounter in schema linking in multilingual environments, mainly stemming from lexical differences (e.g., Japanese hiragana and katakana, Russian Cyrillic characters, and morphological variations in German and French) and the complexity of NoSQL structures (e.g., nested documents and array processing). These factors collectively increase the model's comprehension difficulty, leading to a significant decline in mapping accuracy.

\begin{figure*}[tbp]
    \centering
    \includegraphics[width=1.0\textwidth]{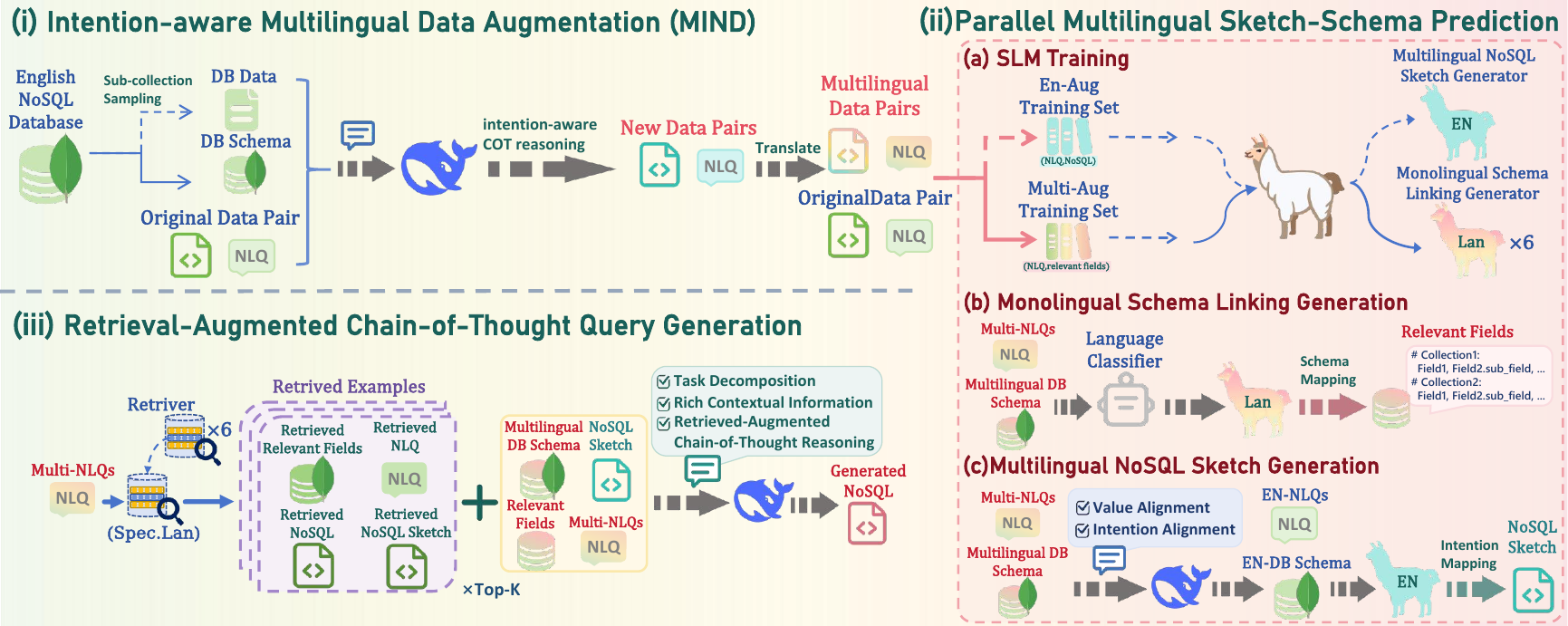}
    \caption{The pipeline of our proposed MultiLink method. MultiLink consists of three main process: (i) Intention-aware Multilingual Data Augmentation (MIND), which enriches training data by generating diverse query pairs through multilingual synthesis and comprehensive database schema analysis; (ii) Parallel Multilingual Sketch-Schema Prediction, which parallelly maps multilingual NLQs intentions to operators and entity mentions to schema elements,including: (a) Multilingual NoSQL Sketch Generation, which generates intermediate NoSQL sketches reflecting operator mappings; (b) Monolingual Schema Linking Generation, which performs precise schema linking for each language;(iii) Retrieval-Augmented Chain-of-Thought Query Prediction, which synthesizes the final NoSQL query by integrating operator and schema mappings with multilingual context.}
    \label{fig:MultiLink_pipeline}
    \vspace{-5pt}
\end{figure*}

\section{Method}
\label{sec Method}
To address the complex challenge of generating NoSQL queries from multilingual NLQs, we introduce the innovative MultiLink framework. 
This framework leverages a problem decomposition strategy and an efficient Parallel Linking Process to effectively address the challenges of multilingual NoSQL query generation. In this chapter, we provide a comprehensive overview of MultiLink.

\subsection{Overview}
\label{sec Method: Overview}
As shown in Figure \ref{fig:MultiLink_pipeline} and Algorithm~\ref{alg:multilink}, MultiLink is primarily divided into three key components: Intention-aware Multilingual Data Augmentation, Parallel Multilingual Sketch-Schema Predictor (including a NoSQL Sketch Generator and a Schema Linking Generator), and Retrieval-Augmented Chain-of-Thought Generator. By leveraging fine-tuned SLM (Small Language Model) technology, which combines low computational resource consumption, shorter training cycles, and sufficient performance to meet our task demands, and integrating it with our data augmentation approach,we achieve cost-effective and high-yield prediction of intention mapping and schema linking. This method specifically targets the extraction of lexical and structural challenges from multilingual NLQs and effectively empowers LLM by providing contextual information, enabling the generation of more accurate and reliable NoSQL queries without significantly exacerbating model hallucinations.
To address these challenges, we employ English, a high-resource language, as a unified bridge for conveying operator information across languages, while utilizing the corresponding languages for schema linking to maintain the model's sensitivity to relevant fields in each language. Finally, through our efficient RAG (Retrieval-Augmented Generation \cite{NEURIPS2020_6b493230}) retrieval technique and a Chain-of-Thought prompting strategy, we consolidate the extracted information into structured contexts. This approach not only mitigates hallucinations in LLMs but also significantly enhances the accuracy of NoSQL generation in multilingual environments.

\subsection{Intention-aware Multilingual Data Augmentation (MIND)}
\label{sec Method: MIND}
We augment the training data using a Intention-aware Chain-of-Thought (CoT) \cite{wei2022chain} guided multilingual query augmentation strategy. Given the original (NLQ, NoSQL) pairs from the TEND dataset, we employ a LLM to synthesize additional pairs with diverse querying intents(for detailed prompt example see~\ref{appendix:prompt in MultiLink}). The augmentation process involves the following steps: (1) analyzing the structural relationships between collections and fields in the MongoDB schema; (2) identifying logical relationships between fields and collections based on the NLQ and the referenced schema portions in the NoSQL query; (3) generating new NoSQL queries with completely different intents from the original queries; (4) creating NLQs that match the intents of the generated NoSQL queries and expanding them into paraphrased variants; and (5) synthesizing corresponding NLQs in multiple languages (e.g., German, French, Russian, Japanese, and Mandarin Chinese). This process not only increases the diversity of multilingual training data but also enhances the performance of the SLMs in intention mapping and schema linking, ensuring that our pipeline can effectively handle multilingual inputs.

\begin{algorithm}[t]
\small
 \SetAlgoLined
 \SetKwFunction{TrainSLM}{\textsc{TrainSLM}}
 \SetKwFunction{BuildVecLib}{\textsc{BuildVecLib}}
 \SetKwFunction{SLMPredict}{\textsc{SLMPredict}}
 \SetKwFunction{Translate}{\textsc{Translate}}
 \SetKwFunction{LangClassify}{\textsc{LangClassify}}
 \SetKwFunction{Retrieve}{\textsc{Retrieve}}
 \SetKwFunction{Generate}{\textsc{Generate}}
 \SetKwFunction{AugmentData}{\textsc{AugmentData}}
 \SetKwFunction{MultiLink}{\textsc{MultiLink}}

 \SetKwInOut{KwIn}{Inputs}
 \SetKwInOut{KwOut}{Output}
 \KwIn{\small{NLQ list in Test Set $\mathcal{Q}$; \\
 DB schema list in Test Set $\mathcal{D}$; \\
 DB list in Training Set $\mathcal{D}'$; \\
 NLQ list in Training Set $\mathcal{Q}'$; \\
 NoSQL list in Training Set $\mathcal{N}'$; \\}}
 \KwOut{\small{NoSQL list $\mathcal{N}$}}
\SetKwProg{Fn}{Procedure}{:}{}
 \Fn{\MultiLink{$\mathcal{Q}$, $\mathcal{D}$}}{
    \tcp{\color{blue} Data Augmentation}
    $\mathcal{Q}'_{\text{aug}}, \mathcal{N}'_{\text{aug}} ,\mathcal{S}'_{\text{aug}} \gets $ \AugmentData{$\mathcal{Q}'$,$\mathcal{N}'$,$\mathcal{D}'$}

    \tcp{\color{blue} Sketch SLM Fine-tuning}
    
    $\mathcal{M}_{\text{n}} \gets $ \TrainSLM{$\texttt{SLM}$, $\mathcal{Q}'_{\text{aug\_en}}$, $\mathcal{N}'_{\text{aug\_en}}$}

    \tcp{\color{blue} Schema Linking SLM Fine-tuning}
    
    $\mathcal{M}_{\text{s}} \gets $ \TrainSLM{$\texttt{SLM}$, $\mathcal{Q}'_{\text{aug}}$, $\mathcal{S}'_{\text{aug}}$}

    \tcp{\color{blue} Build Vector Library}

    $\mathcal{V} \gets $ \BuildVecLib{$\mathcal{Q}'$, $\mathcal{N}'$}

    \tcp{\color{blue} Pipeline of MultiLink}
    
    $\mathcal{N} \gets \texttt{[]}$
    
    \For{\textup{\textbf{each} $(q, d) \in (\mathcal{Q}, \mathcal{D})$}}{
        \tcp{\color{purple} Language Classification}
        $L$ $\gets$ \LangClassify{$q$}

        \tcp{\color{purple} Translation}
        $q_{\text{en}}$, $d_{\text{en}}$ $\gets$ \Translate{$q$, $d$}
        
        \tcp{\color{purple} Sketch Prediction}
        $n_{\text{sk}}$ $\gets$ \SLMPredict{$\mathcal{M}_{\text{n}}$, $q_{\text{en}}$, $d_{\text{en}}$}

        \tcp{\color{purple} Schema Linking Prediction}
        $s$ $\gets$ \SLMPredict{$\mathcal{M}_{\text{s}}$, $q$, $d$}

        \tcp{\color{purple} Retrieval-Aug CoT Generation}
        $\mathcal{E}$ $\gets$ \Retrieve{$q$,$\mathcal{V}$,$L$}

        $n_{\text{gen}}$ $\gets$ \Generate{$q$, $d$,$n_{\text{sk}}$, $s$,$\mathcal{E}$}

        $\mathcal{N}$.append($n_{\text{gen}}$)
    }
    \KwRet $\mathcal{N}$
  }
 \caption{The MultiLink Algorithm}
 \label{alg:multilink}
\end{algorithm}

\subsection{Parallel Multilingual Sketch-Schema Predictor}
The Parallel Multilingual Sketch-Schema Predictor is a key component of our pipeline, designed to address the lexical and structural challenges in multilingual NoSQL generation in parallel. The predictor consists of two parallel submodules: (i) the Multilingual NoSQL Sketch Generator, which maps multilingual NLQs intents to NoSQL operators via a unified intermediate representation (i.e., English); and (ii) the Monolingual Schema Linking Generator , which maps entity mentions in the NLQ to the corresponding schema elements in the target database. By executing these submodules in parallel, Sketch-Schema Predictor ensures high accuracy and efficiency in both operator and schema mapping across multilingual contexts.

\paragraph{Multilingual NoSQL Sketch Generator} 
To address the challenge in intention mapping exacerbated by lexical diversity and syntactic heterogeneity in multilingual contexts, we designed Multilingual NoSQL Sketch Generator, a sketch generator incorporating the mapping from intention to operator. We adopt English, a high-resource language, as a unified bridge for cross-lingual transfer of operator information. Given a multilingual NLQ, Sketch Generator, leveraging the power of an LLM, extracts and anchors the underlying query intent by translating both the NLQ and the database schema into English. The translated English NLQ, along with the corresponding database schema, is then fed into the fine-tuned SLM to generate an intermediate NoSQL sketch. This sketch reflects operator mappings (e.g., sort, filter, aunwind), but does not include precise schema field references. By unifying multilingual intentions into a single language (i.e., English), Sketch Generator efficiently and cost-effectively streamlines the operator mapping process and ensures consistency across languages.

\paragraph{Monolingual Schema Linking Generator} 
\begin{CJK*}{UTF8}{gbsn}
In the Multilingual Text-to-NoSQL task, models are typically required to have a thorough understanding of entity mentions across complex lexicons in different languages (e.g., hiragana and katakana in Japanese, Cyrillic characters in Russian, and the rich and varied lexical forms in German and French). At the same time, they must possess the ability to cross-linguistically map entity mentions in NLQ to the corresponding fields in the database schema. The nested and unstructured nature of NoSQL schemas further exacerbates the challenge in schema linking.
Therefore, we designed an efficient format to express schema linking results, such as `\# Collection1: Field1, Field2.sub\_field,... \textbackslash n \# Collection2:..' (e.g., ‘\# 产品: 产品价格, 投诉.员工ID\textbackslash n\# 员工: 员工ID’). Based on this format, we constructed corresponding corpora for schema linking in different languages. Combined with a language classifier, the schema linking generator inputs the multilingual NLQs (e.g., ``登山者为位于乌干达的山峰记录的攀登时间是什么？'') into a fine-tuned SLM (Schema Linking Model) trained on language-specific schema linking corpora, in order to accurately map the entity mentions in the NLQ to the corresponding schema elements in the target database(e.g., \# 山脉: 国家, 登山者, 登山者.时间). By employing separately fine-tuned SLMs for schema linking in each language, Schema Linking Generator ensures high accuracy in schema linking result, addressing the lexical challenges in multilingual NoSQL generation.
\end{CJK*}

\subsection{Retrieval-Augmented Chain-of-Thought Query Generator}
The final module of our pipeline is the Retrieval-Augmented Chain-of-Thought Generator, which synthesizes the final NoSQL query by integrating the results from the previous steps. Given a multilingual NLQ, the Query Generator include: (i) the reference English NoSQL query with operator mappings (from Sketch Generator); (ii) the database schemas; (iii) the schema linking result of current NLQ (from Schema Linking Generator); and (iv) retrieved examples from the corresponding language example library created from the training data.
Using a Retrieval-Augmented Chain-of-Thought reasoning approach, the Query Generator significantly enhances the reasoning capabilities of the LLM by referencing similar retrieved examples in the same language and guiding the query generation process step-by-step. By combining the results from Sketch Generator and Schema Linking Generator, the Query Generator addresses the inherent challenges of multilingual scenarios. Leveraging the enhanced reasoning capabilities of the LLM, Query Generator accurately synthesizes and utilizes key contextual information, generating precise and semantically consistent NoSQL queries with higher scores and better performance compared to baseline models.

\begin{table}[th!]
  \centering
  \begin{subtable}[t]{0.45\textwidth}
    \centering
    \begin{tabular}{lccc}
      \toprule
      Model & EM & QSM & QFC \\
      \midrule
      Fine-tuned Llama & 15.64\% & 55.90\% & 58.44\% \\
      Zero-shot LLM & 0.48\% & 49.35\% & 59.36\% \\
      Few-shot LLM & 10.82\% & 56.17\% & 61.12\% \\
      RAG for LLM & 14.32\% & 59.79\% & 67.19\% \\
      SMART & 17.23\% & 59.12\% & 61.94\% \\
      \textbf{MultiLink (Ours)} & \color{high_score}\textbf{25.54\%} & \color{high_score}\textbf{64.01\%} & \color{high_score}\textbf{73.17\%} \\
      \bottomrule
    \end{tabular}
    \caption{Query-based Metric Results (Avg of 6 langs)}
    \label{subtab:1}
  \end{subtable}
  \hfill 
  \begin{subtable}[t]{0.45\textwidth}
    \centering
    \begin{tabular}{lccc}
      \toprule
      Model & EX & EFM & EVM \\
      \midrule
      Fine-tuned Llama & 40.95\% & 81.41\% & 70.46\% \\
      Zero-shot LLM & 32.44\% & 55.98\% & 59.22\% \\
      Few-shot LLM & 36.69\% & 64.43\% & 64.71\% \\
      RAG for LLM & 47.95\% & 74.22\% & 69.80\% \\
      SMART & 43.36\% & 85.05\% & \color{high_score}\textbf{76.37\%} \\
      \textbf{MultiLink (Ours)} & \color{high_score}\textbf{59.12\%} & \color{high_score}\textbf{85.66\%} & 74.01\% \\
      \bottomrule
    \end{tabular}
    \vspace{-5pt}
    \caption{Execution-based Metric Results (Avg of 6 langs)}

    \label{subtab:2}
  \end{subtable}
  \caption{Overall Performance Metrics}
  \label{tab:Performance Comparison}
\end{table}

\section{Experiments and Analysis}

\subsection{Experimental Setup}
\paragraph{Dataset}
We conducted cross-domain partitioning of the MultiTEND dataset for different languages, ensuring that the training set for each language contained the same sample content. The dataset was divided into original language-specific training and test sets at a ratio of 0.85:0.15, and the training and test sets for each language were merged to form a multilingual training set and test set encompassing six languages. For the additional dataset obtained through data augmentation, which contains 2,666 different NoSQL queries with 5 NLQs corresponding to each NoSQL query in each language, we directly added it to the original training set to create an augmented language-specific training set. By combining these datasets for each language, we obtained a multi-augmented training set covering six languages. Comprehensive statistics on dataset splits, including distinct training sets used across MultiLink modules, are provided in the Appendix \ref{appendix: Dataset Details}.

\paragraph{Baselines} 
We utilized a variety of popular neural network models (i.e., LLM-based prompting methods (i.e., \textbf{Zero-shot LLM}, \textbf{Few-shot LLM}, \textbf{RAG for LLM}), SLM-based fine-tuning methods (i.e., \textbf{Fine-tuned SLM}) and existing Text-to-NoSQL methods (i.e., \textbf{SMART} \cite{lu2024natural}) as baseline models for a comprehensive performance comparison with MultiLink. 
The details of these baseline models can be found in Section~\ref{baseline}
 of the Appendix.

\paragraph{Evaluation Metrics} Following other text-to-NoSQL study like SMART \cite{lu2024natural}, we report results using the same metrics including \textit{Exact Match} (EM) and \textit{Execution Accuracy} (EX), each with more detailed subdivisions such as \textit{Query Stages Match} (QSM) and \textit{Query Fields Coverage} (QFC) under EM, and \textit{Execution Fields Match} (EFM) and \textit{Execution Value Match} (EVM) under EX.
The detailed definition of these metrics can be found in Section~\ref{metrics} of the Appendix. 

\paragraph{Implementation Details}

The SLM used in MultiLink is ``Llama-3.2-1B'', with a full-parameter fine-tuning strategy and set to 3 epochs. The LLM used is ``DeepSeek-V3'', with the parameter setting `temperature = 0.0'. The text-to-embedding model used is ``text-embedding-ada-002''.

\subsection{Performance Comparison}
Table \ref{tab:Performance Comparison} presents the average performance of MultiLink and baseline methods across all languages in MultiTEND (For detailed per-language and per-metric analysis of MultiLink and all baselines, please refer to the Appendix~\ref{appendix: Performance Comparison}). As shown in Table~\ref{tab:Performance Comparison}, the fine-tuned Llama~\cite{dubey2024llama} and LLM-based methods (Zero/Few-shot LLM, RAG for LLM) underperform below 50\%, with Zero-shot LLM systematically exhibiting the weakest results due to deficiencies in query intent comprehension and critical failures in processing nested arrays/multi-set associations. While the approach of RAG for LLM with enhanced contextual information shows relatively better performance in the Execution Accuracy (EX) metric that directly reflects query execution outcomes. This suggests that neither pure fine-tuning nor direct reliance on LLM capabilities constitutes an effective solution for multilingual NoSQL challenges.

Additionally, while SMART, which is designed for English contexts, performs averagely in multilingual NoSQL tasks, showing significantly different results between English and non-English languages (see Table \ref{tab:Performence Comparison of baselines on MultiTEND}). This indicates that existing Text-to-NoSQL systems, cannot be directly extended to non-English scenarios. In contrast, our proposed MultiLink framework demonstrates superior performance in multilingual environments, outperforming all existing models across every metric. Particularly noteworthy is its 11\% absolute improvement over the best-performing baseline in the crucial EX metric. These results validate that MultiLink's design effectively addresses multilingual NoSQL generation challenges and produces high-quality queries.Due to space limitations, please refer to appendix \ref{appendix: Performance Comparison} for more detailed analysis.

\begin{figure}[t!]
    \centering
    \begin{subfigure}[b]{0.45\textwidth}
        \centering
        \includegraphics[width=\textwidth]{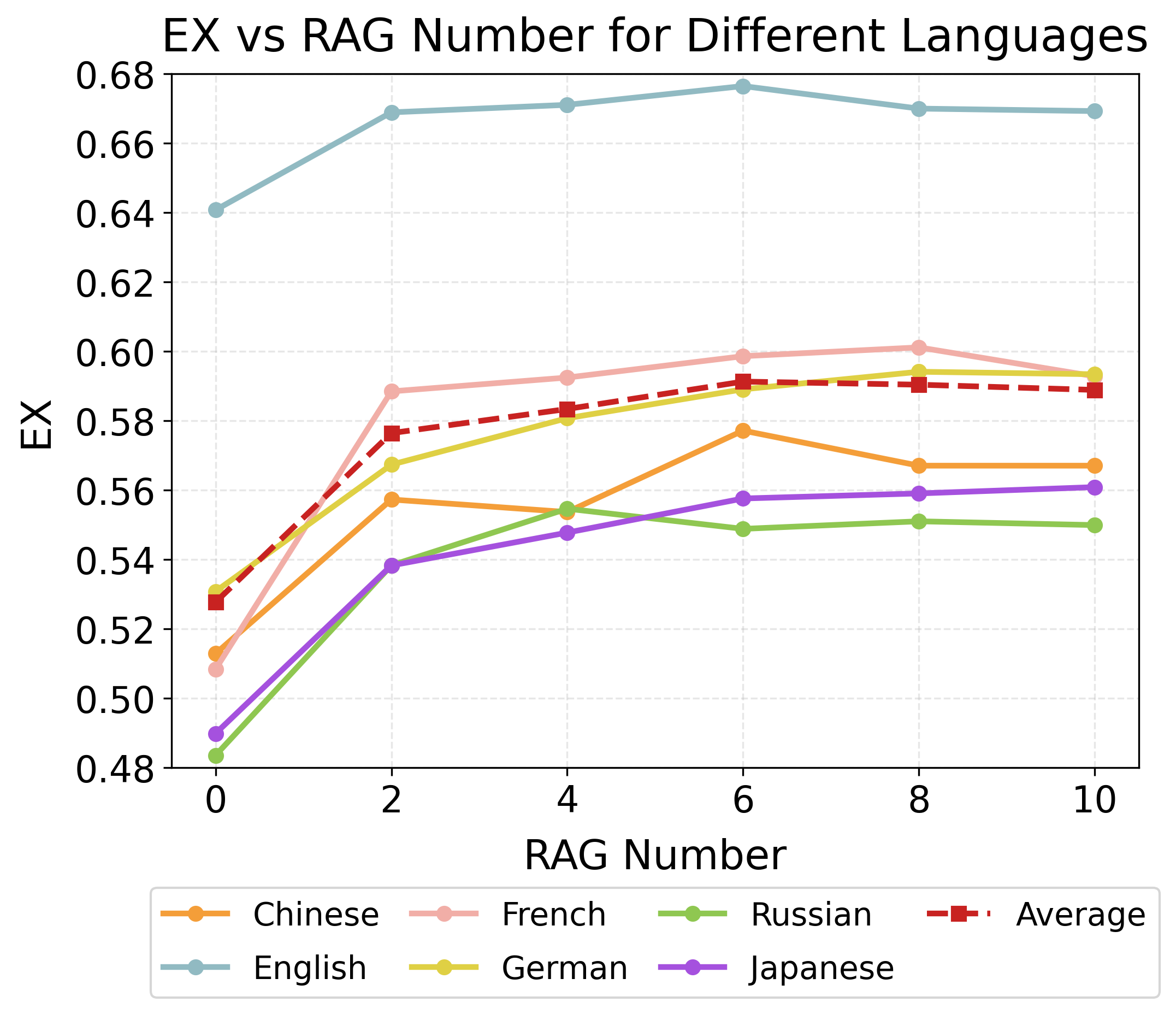}
        \caption{Execution-based Metric}
    \end{subfigure}
    \caption{{Parameter study}}
    \label{Figure: Parameter Study}
\end{figure}

\begin{table}[th!]
  \centering
  \small 
  \begin{subtable}[t]{0.45\textwidth}
    \centering
    \begin{tabular}{lccc}
      \toprule
      \textbf{Method} & EM & QSM & QFC \\
      \midrule
      Few-Shot LLM & 10.82\% & 56.17\% & 61.12\% \\
      RAG for LLM & 14.32\% & 59.79\% & 67.19\% \\
      \textbf{MultiLink (Ours)} & 25.54\% & 64.01\% & 73.17\% \\
      $\quad$ - w/o Sk-G & 25.51\% & 63.97\% & 73.12\% \\
      $\quad$ - w/o SL-G & \textbf{25.55\%} & \textbf{64.06\%} & \textbf{73.19\%} \\
      $\quad$ - w/o AUG  & 21.18\% & 61.46\% & 70.15\% \\
      $\quad$ - only GEN & 14.40\% & 61.68\% & 67.51\% \\
      \bottomrule
    \end{tabular}
    \caption{Query-based Metric}
    \label{subtab:query}
  \end{subtable}
  \hfill
  \begin{subtable}[t]{0.45\textwidth}
    \centering
    \begin{tabular}{lccc}
      \toprule
      \textbf{Method} & EX & EFM & EVM \\
      \midrule
      Few-Shot LLM & 36.69\% & 64.43\% & 64.71\% \\
      RAG for LLM & 47.95\% & 74.22\% & 69.80\% \\
      \textbf{MultiLink (Ours)} & \textbf{59.12\%} & 85.66\% & \textbf{74.01\%} \\
      $\quad$ - w/o Sk-G & 58.82\% & 85.64\% & 73.75\% \\
      $\quad$ - w/o SL-G & 59.03\% & \textbf{85.69\%} & 73.83\% \\
      $\quad$ - w/o AUG  & 55.74\% & 84.48\% & 73.36\% \\
      $\quad$ - only GEN & 47.19\% & 73.33\% & 69.71\% \\
      \bottomrule
    \end{tabular}
    \caption{Execution-based Metric}
    \label{subtab:execution}
  \end{subtable}
  \caption{Ablation study results on MultiTEND.}
  \label{tab:ablation_study}
\end{table}

\subsection{Parameter Study}
To explore the performance of UniLink under different parameters, we conducted a hyperparameter experiment on the number of retrieved examples (RAG num) using the test set of the MultiTEND dataset. As shown in Figure \ref{Figure: Parameter Study}, the figures illustrate the execution accuracy (EX) of MultiLink under different RAG numbers across multiple languages (for complete metric results of the parameter study, see the appendix \ref{appendix: Parameter Study}). As the RAG number increases, MultiLink exhibits slight fluctuations in various metrics across different languages. The execution accuracy (EX) shows that English performs the best, with minor fluctuations around 67\%; French and German are in the middle range, at 58\-60\%; while Chinese, Japanese, and Russian are lower, consistently staying within the 54\%-57\% range. The average performance across all languages (represented by the red dashed line) shows a steady increase and stabilizes after reaching a RAG number of 6. This indicates that as the RAG number increases, the model's performance improves within a certain range, and MultiTEND achieves its best performance at a RAG number of 6.

\subsection{Ablation Study}
In this section, we conduct an ablation study to examine the contribution of each module in MultiLink. We first measure the results of MultiLink based on the complete pipeline designed. Then, we evaluate the contribution of each module by removing different key modules from MultiLink, specifically: (i) removing the Sketch Generator (w/o Sk-G); (ii) removing the Schema Linking Generator (w/o SL-G); (iii) using only the Retrieval-Augmented Chain-of-Thought Generator (only GEN); and (iv) using components without data augmentation (w/o AUG).

Additionally, we include the experimental results of Few-shot LLM and Retrieval-Augmented Generation (RAG) for LLM to compare with the ablation study results. This comparison aims to demonstrate that the high performance of MultiLink is not solely reliant on the inherent capabilities of the LLM itself, but rather stems from our designed complex and effective pipeline.

The results of the ablation experiments are shown in Table \ref{tab:ablation_study}. MultiLink with all processes included outperforms other configurations across all metrics to varying degrees. Among them, the performance of w/o Sk-G is relatively close to that of MultiLink with all processes included, while w/o SL-G demonstrates that the contextual information provided by the SL-G module is crucial, significantly aiding the LLM in generating more accurate queries. The results of w/o AUG are the lowest, proving that our data augmentation method substantially enhances the performance of each module in the model. Overall, on the EX metric, which best reflects the model's performance in real-world scenarios, MultiLink with all processes included outperforms all other configurations. This validates the effective contribution of all components in MultiLink to the overall framework.

Furthermore, on the critical EX and EM metrics, MultiLink with all major processes included significantly outperforms Few-shot LLM, RAG for LLM, only GEN, and w/o AUG configurations. This indicates that the high accuracy of MultiLink does not directly stem from the inherent understanding and generation capabilities of the LLM itself, but rather primarily from the framework itself and the information provided by the SLM enhanced using our designed data augmentation method.

\section{Conclusion}

In this work, we introduce MultiTEND, a large-scale multilingual benchmark dataset for Text-to-NoSQL tasks encompassing six languages. To create this dataset, we developed a robust process that combines the capabilities of LLMs with human efforts. This approach ensures high-quality, semantically aligned, and contextually accurate database fields, NLQs, and NoSQL queries through thorough manual verification.
Next, we identify the inherent challenges of multilingual Text-to-NoSQL tasks, including lexical variations and structural inconsistencies across languages. To address these issues, we propose MultiLink, a unified multilingual pipeline that breaks down the complex task into manageable steps, such as multilingual query augmentation and language-specific schema linking.
Extensive experiments demonstrate that MultiLink excels in generating accurate and semantically consistent NoSQL queries across multiple languages, significantly outperforming existing baseline models.

Building on this line of research, we aim to explore additional methodologies for text-to-NoSQL tasks as the next phase of our work. 
We anticipate that this work will not only contribute to the ongoing evolution of the NoSQL field but also inspire further innovations, fostering a dynamic research landscape similar to the advancements seen in the parallel text-to-SQL domain.

\section{Limitation}
We propose a unified multilingual Text-to-NoSQL pipeline, which effectively addresses the lexical and structural challenges in multilingual NoSQL generation by integrating context information generated from fine-tuned SLMs and adopting a multi-step approach that combines CoT and RAG prompting methods. Additionally, our designed data augmentation method further enhances the accuracy and quality of NoSQL query generation by the framework.
However, our research in multilingual aspects is still limited to six languages (English, German, French, Russian, Japanese, and Mandarin Chinese), which only cover a portion of the mainstream languages within the Indo-European and Sino-Tibetan language families, while neglecting the needs of other language families.
The experimental results are constrained by the limited scope of general-purpose LLMs. For instance, although we use relatively advanced and high-performance LLMs in our experiments, there is a lack of exploration into methods that could enable lower-performance but more cost-efficient LLMs to achieve similar results on this task.
The pipeline requires high computational costs for LLMs. For example, in scenarios where LLMs are used in the pipeline, for obtaining higher-quality outputs, the long-context inputs with rich examples and step-by-step reasoning outputs, significantly increases token overhead.
Therefore, future research could expand to include more widely used languages, explore the application of Text-to-NoSQL in low-resource or minority languages, and investigate the use of other LLM architectures or the development of more cost-effective and high-performance neural-based framework strategies.


\bibliography{acl_latex}
\onecolumn
\newpage
\appendix

\section{Related Work}
This study is closely related to the fields of Text-to-SQL and NoSQL Databases, as briefly surveyed below.
\subsection{Text-to-SQL}
Early research on Text-to-SQL primarily focused on meticulously designed rule-based methods, such as those in \citep{baik2020duoquest,li2014constructing,li2014nalir,quamar2022natural,sen2020athena++}, these methods used predefined rules or semantic parsers to translate NLQs into SQL but were inflexible and inadequate for handling increasingly complex database structures.
With the rise of deep learning, the focus of Text-to-SQL research has gradually shifted towards methods that utilize deep neural networks, such as attention mechanisms~\citep{liu2023multi}  and graph-based encoding strategies~\citep{hui-etal-2022-s2sql,li2023graphix,qi-etal-2022-rasat,wang-etal-2020-rat,xu-etal-2018-sql,zheng-etal-2022-hie,yu2021grappa,xiang2023g3r}. Alternatively, some approaches treat Text-to-SQL as a sequence-to-sequence problem by using encoder-decoder structured Pre-trained Language Models (PLMs) to generate SQL queries~\citep{10.5555/3304222.3304323,popescu2022addressing,qi-etal-2022-rasat}.

In recent years, large language models (LLMs), which have demonstrated remarkable success across various domains, have also garnered increasing attention in the Text-to-SQL field~\citep{dong2023c3,gan-etal-2021-natural-sql,10.14778/3641204.3641221,li2023resdsql,lin-etal-2020-bridging,pourreza2024din,qi-etal-2022-rasat,rubin-berant-2021-smbop,scholak-etal-2021-picard}. Current literature primarily focuses on two approaches with LLMs: prompt engineering and pretraining/fine-tuning. Prompt engineering methods typically involve using specific reasoning workflows which can be categorized into several reasoning modes, including Chain-of-Thought (CoT) \citep{wei2022chain} and its variants \citep{pourreza2024din,liu2023divide,zhang-etal-2024-coe,zhang-etal-2023-act}, Least-to-Most \citep{zhou2023leasttomost,gan-etal-2021-natural-sql,arora-etal-2023-adapt}, and Decomposition \citep{khot2023decomposed,tai-etal-2023-exploring,pourreza2024din,wang-etal-2025-mac,xie-etal-2024-decomposition}. To evaluate Text-to-SQL model performance in practical applications, several large-scale benchmark datasets have been developed and released, including WikiSQL \citep{zhong2018seqsql}, Spider \citep{yu-etal-2018-spider}, KaggleDBQA \citep{lee-etal-2021-kaggledbqa}, BIRD \citep{10.5555/3666122.3667957}, and Bull \citep{10.1145/3626246.3653375} etc.

\subsection{NoSQL Database}

Traditional SQL databases face limitations with large-scale, unstructured, or semi-structured data in the internet and big data era, prompting the rise of NoSQL databases, which provide flexibility, scalability, and high performance in web applications and real-time data analysis \cite{moniruzzaman2013nosql}.
In the field of databases and NLP, current research primarily focuses on several key areas of NoSQL databases,including achieving scalability in data storage systems within large-scale user environments \citep{cattell2011scalable}, ensuring consistency in NoSQL databases \citep{diogo2019consistency}, addressing multi-tenant NoSQL data storage issues in cloud computing environments, particularly in scenarios involving resource and data sharing \citep{zeng2015resource}, and realizing scalability, elasticity, and autonomy in database management systems (DBMS) within cloud computing environments \citep{agrawal2011database}.

Despite the extensive research on NoSQL across various domains, its accessibility remains a challenge, especially for non-expert users. Although Text-to-NoSQL tasks have been proposed to address this issue, existing NoSQL generation primarily supports English and overlooks the needs of non-English users.
To tackle this issue, we introduce the Multilingual Text-to-NoSQL task, which is based on existing Text-to-NoSQL research and not only aims to reduce the barrier for non-expert users to utilize NoSQL databases by automatically converting NLQs into NoSQL queries but also addresses the gap in existing Text-to-NoSQL tasks that mainly support English while neglecting non-English users' needs. In this task, we also introduce MultiTEND, the largest multilingual benchmark for natural language to NoSQL query generation.

\newpage

\section{Dataset Analysis}

\subsection{Dataset Details}
\label{appendix: Dataset Details}

\begin{figure*}[h!]
    \centering
    \begin{subfigure}[b]{0.25\textwidth}
        \centering
        \includegraphics[width=\textwidth]{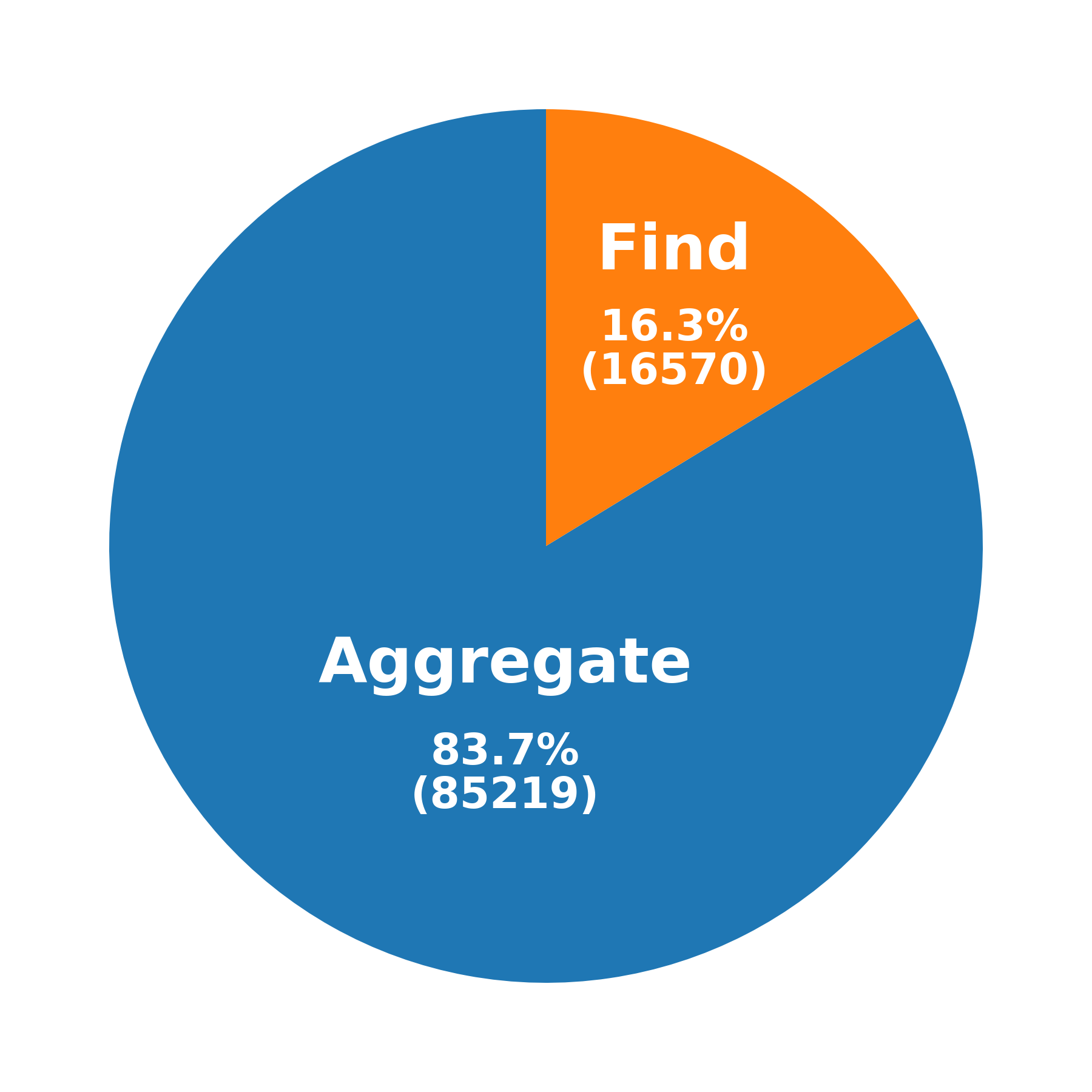}
        \caption{NoSQL Query Method}
        \label{appx.fig:NoSQL Query Method}
    \end{subfigure}
    \hfill
    \begin{subfigure}[b]{0.375\textwidth}
        \centering
        \includegraphics[width=\textwidth]{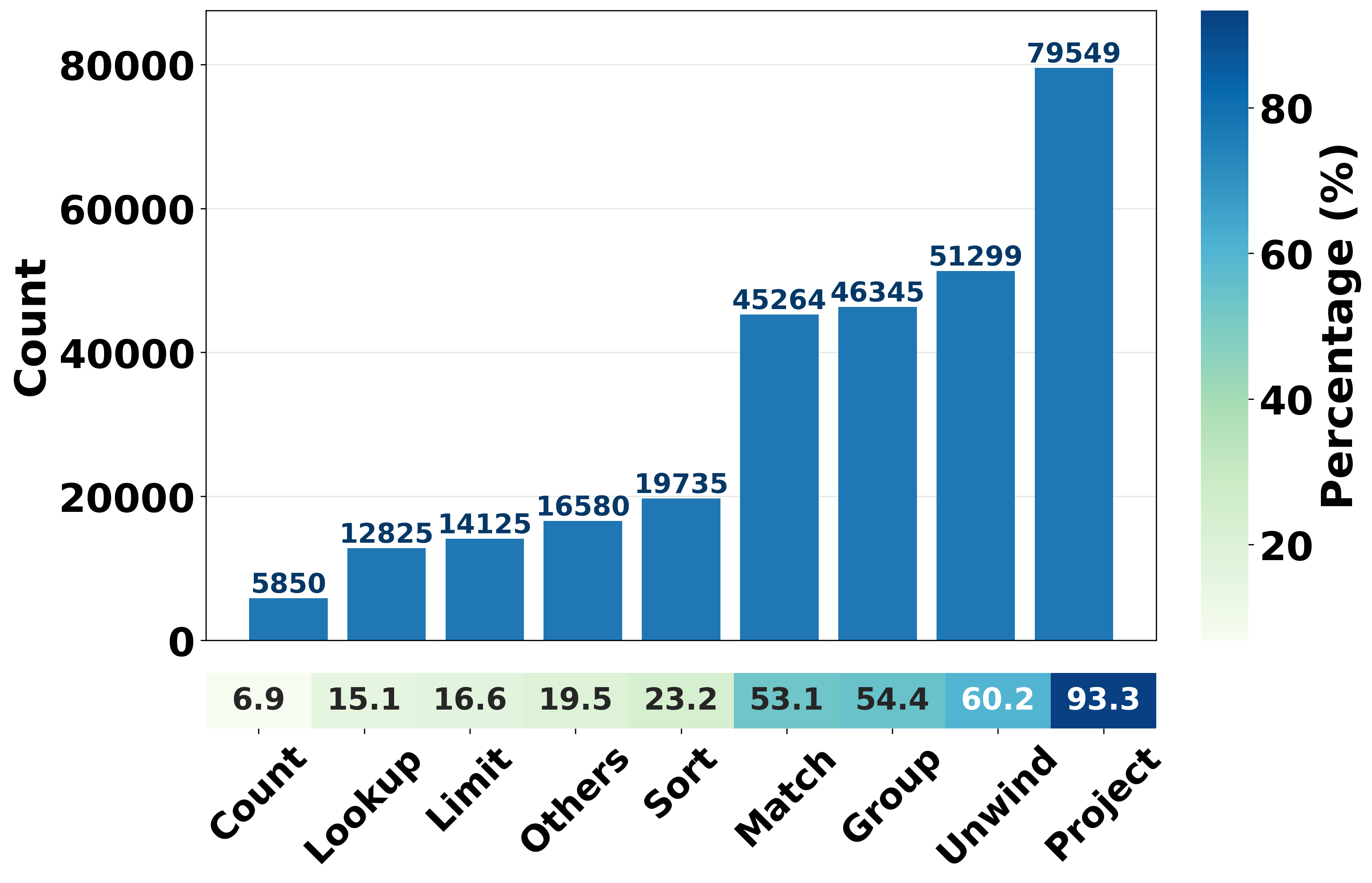}
        \caption{Stages in Aggregate Method}
        \label{appx.fig:Stages in Aggregate Method}
    \end{subfigure}
    \hfill
    \begin{subfigure}[b]{0.355\textwidth}
        \centering
        \includegraphics[width=\textwidth]{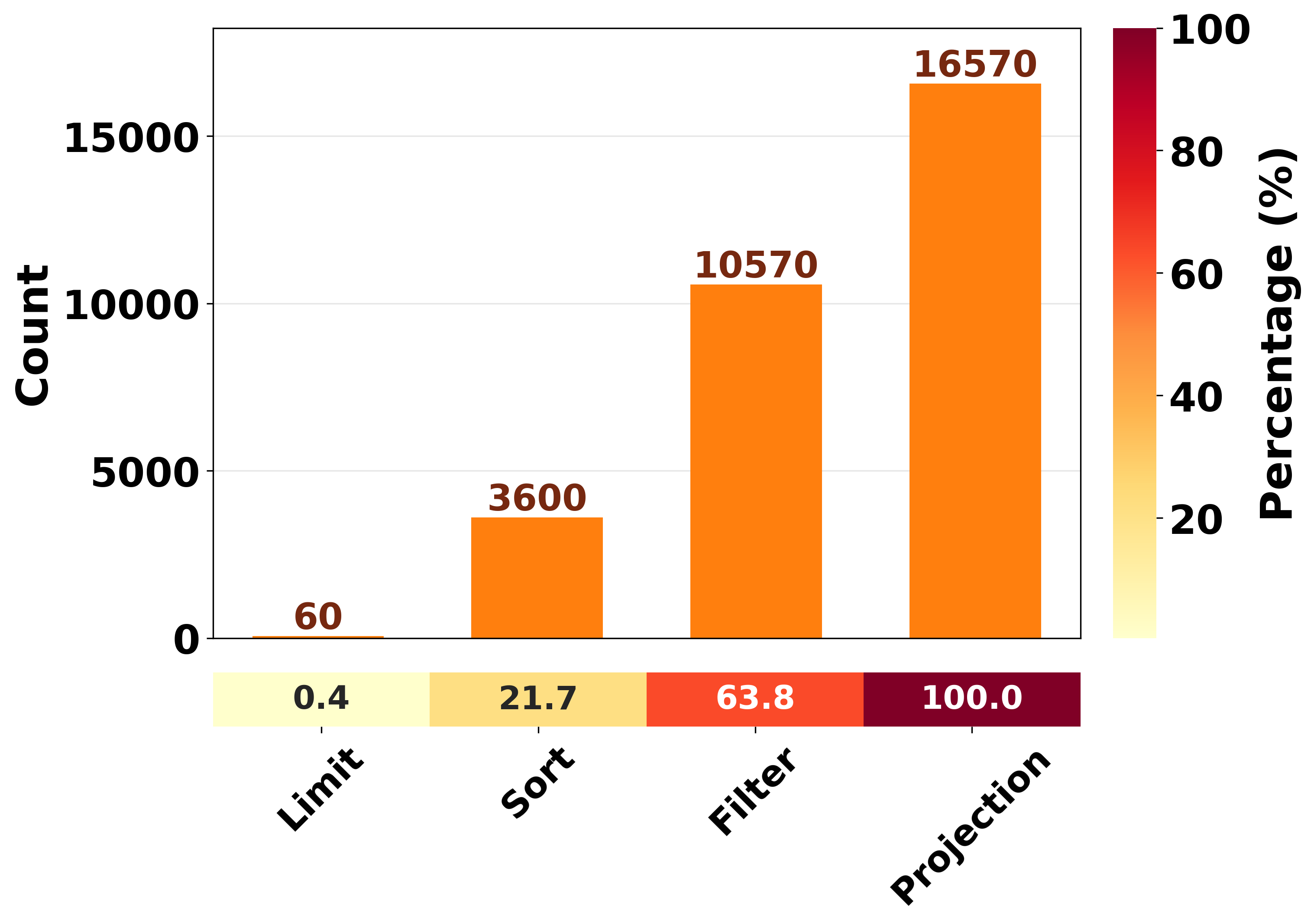}
        \vspace{-0.6cm}
        \caption{Operations in Find Method}
        \label{appx.fig:Operations in Find Method}
    \end{subfigure}
    \caption{NoSQL Query Statistics in MultiTEND}
    \label{appx.fig:NoSQL Query Statistics of MultiTEND}
\end{figure*}

\begin{table}[h!]
\centering
\begin{minipage}{0.3\textwidth} 
    \centering
    \includegraphics[width=\linewidth]{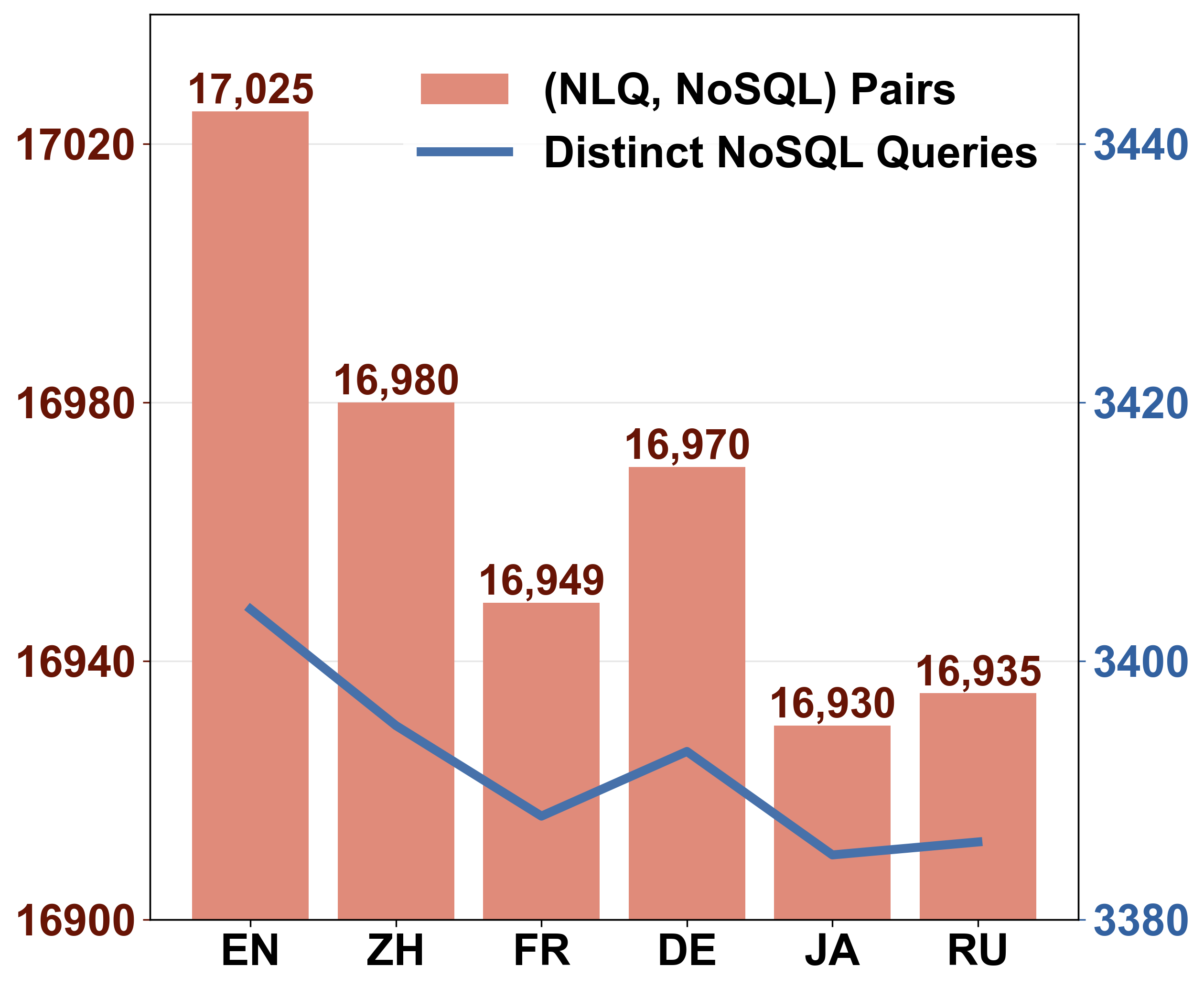}
    \vspace{-16.5pt}
    \captionof{figure}{(NLQ,Query) Statistics} 
    \label{appx.fig:(NLQ,Query) Statistics}
\end{minipage}
\hspace{20pt}
\begin{minipage}{0.60\textwidth} 
    \centering
    \begin{tabular}{lllll}
        \rowcolor[HTML]{DD7E6B} 
        \textcolor[HTML]{FFFFFF}{\textbf{\#-Database}} & 
        \textcolor[HTML]{FFFFFF}{\textbf{\#-Domain}} & 
        \textcolor[HTML]{FFFFFF}{\textbf{\#-Collections}} & 
        \textcolor[HTML]{FFFFFF}{\textbf{\#-Documents}} & 
        \textcolor[HTML]{FFFFFF}{\textbf{\#-Fields}} \\
        
        924 & 105 & 2082 & 197874 & 35760 \\
        
        \rowcolor[HTML]{DD7E6B} 
        \multicolumn{5}{c}{\textcolor[HTML]{FFFFFF}{\textit{\textbf{Top-5 Domains}}}} \\
        
        Sport & Customer & School & Shop & Student \\
        
        \rowcolor[HTML]{FFFFFF} 
        
        \rowcolor[HTML]{DD7E6B} 
        \multicolumn{5}{c}{\color[HTML]{FFFFFF} \textit{\textbf{Statistics of Database}}} \\
        
        \rowcolor[HTML]{DD7E6B} 
        & 
        {\color[HTML]{FFFFFF} \textbf{\#-Avg}} & 
        {\color[HTML]{FFFFFF} \textbf{\#-Max}} & 
        {\color[HTML]{FFFFFF} \textbf{\#-Min}} & 
        {\color[HTML]{FFFFFF} \textbf{\#-Total}} \\
        
        \cellcolor[HTML]{DD7E6B}{\color[HTML]{FFFFFF} \textbf{Cols}}   & 2.25  & 7     & 1    & {\color[HTML]{000000} 2082} \\
        \cellcolor[HTML]{DD7E6B}{\color[HTML]{FFFFFF} \textbf{Docs}}   & 214.15 & 13694 & 3    & 197874 \\
        \cellcolor[HTML]{DD7E6B}{\color[HTML]{FFFFFF} \textbf{Fields}} & 38.7   & 331   & 7    & 35760 \\
    \end{tabular}
    \caption{Database Statistics in MultiTEND}
    \label{tab:MultiTEND dataset statistics}
\end{minipage}
\end{table}
Figure \ref{appx.fig:(NLQ,Query) Statistics} presents detailed statistics of (NLQ, NoSQL) Pairs and distinct NoSQL Queries across different languages in MultiTEND. Figure \ref{appx.fig:NoSQL Query Statistics of MultiTEND} displays the statistics of NoSQL queries in MultiTEND (covering all six languages). Specifically, figure \ref{appx.fig:NoSQL Query Method} uses a pie chart to illustrate the distribution of different query methods in MultiTEND, while figure \ref{appx.fig:Stages in Aggregate Method} and figure \ref{appx.fig:Operations in Find Method} show the counts of stages in the aggregate method and operators in the find method, respectively, with a heatmap included to represent the proportion of queries that use each specific stage or operator relative to the total number of queries employing the corresponding method. 
\par Table \ref{tab:MultiTEND dataset statistics} conveys detailed statistical information about all databases (covering six languages) in the MultiTEND dataset, which includes a total of 924 databases spanning 105 domains. The top five most represented domains in the dataset are Sport, Customer, School, Shop, and Student. Across all databases, there are a total of 2,082 collections, 197,874 documents, and 35,760 fields.
Each database contains an average of 2.25 collections, with values ranging from 1 to 7. The number of documents averages 214.15 per database, spanning from 3 to 13,694. Field counts average 38.7 per database, varying from 7 to 331.
\newpage
\subsection{Analysis and Findings}
\label{appendix: Analysis and Findings}
\begin{table*}[th!]
\centering
{
\begin{tabular}{llccccccc}
\toprule
\rowcolor{row_color}
\textbf{Metric} & \textbf{Model} & \textbf{EN} & \textbf{ZH} & \textbf{FR} & \textbf{DE} & \textbf{JA} & \textbf{RU} & \textbf{AVG (5 langs)} \\
\midrule
\multirow{4}{*}{EM} 
 & Fine-tuned Llama & 17.05\% & 13.57\% & 16.53\% & 15.78\% & 16.40\% & 14.51\% & 15.36\% \\
 & Zero-shot LLM & \color{low_score}0.29\% & 0.61\% & 0.61\% & 0.54\% & 0.54\% & 0.29\% & \color{low_score}0.52\% \\
 & RAG for LLM & 16.09\% & 13.98\% & 15.62\% & 14.33\% & 12.02\% & 13.89\% & 13.97\% \\
 & SMART & \color{high_score}18.85\% & 13.94\% & 18.38\% & 18.30\% & 18.05\% & 15.89\% & \color{high_score}16.91\% \\
\cmidrule{1-9}
\multirow{4}{*}{QSM} 
 & Fine-tuned Llama & 57.19\% & 56.71\% & 54.22\% & 56.14\% & 56.22\% & 54.91\% & 55.64\% \\
 & Zero-shot LLM & \color{low_score}51.24\% & 47.76\% & 50.36\% & 50.43\% & 47.35\% & 48.95\% & \color{low_score}48.97\% \\
 & RAG for LLM & \color{high_score}62.30\% & 59.52\% & 60.51\% & 60.17\% & 57.36\% & 58.87\% & \color{high_score}59.28\% \\
 & SMART & 61.15\% & 57.69\% & 61.23\% & 59.35\% & 58.11\% & 57.17\% & 58.71\% \\
\cmidrule{1-9}
\multirow{4}{*}{QFC} 
 & Fine-tuned Llama & 60.76\% & 53.83\% & 56.61\% & 62.35\% & 58.70\% & 58.41\% & \color{low_score}57.98\% \\
 & Zero-shot LLM & \color{low_score}60.29\% & 58.59\% & 58.92\% & 60.07\% & 58.09\% & 60.22\% & 59.18\% \\
 & RAG for LLM & \color{high_score}68.04\% & 67.86\% & 67.03\% & 67.47\% & 65.29\% & 67.46\% & \color{high_score}67.02\% \\
 & SMART & 65.05\% & 60.36\% & 60.97\% & 63.86\% & 62.03\% & 59.34\% & 61.31\% \\
\cmidrule{1-9}
\multirow{4}{*}{EX} 
 & Fine-tuned Llama & 44.61\% & 36.86\% & 41.26\% & 41.44\% & 43.32\% & 38.23\% & 40.22\% \\
 & Zero-shot LLM & \color{low_score}36.58\% & 28.99\% & 33.86\% & 34.91\% & 30.63\% & 29.68\% & \color{low_score}31.61\% \\
 & RAG for LLM & \color{high_score}51.70\% & 47.02\% & 49.28\% & 48.59\% & 45.12\% & 45.99\% & \color{high_score}47.20\% \\
 & SMART & 48.86\% & 38.05\% & 44.69\% & 44.22\% & 43.30\% & 41.03\% & 42.26\% \\
\cmidrule{1-9}
\multirow{4}{*}{EFM} 
 & Fine-tuned Llama & 84.97\% & 78.84\% & 80.14\% & 81.44\% & 79.50\% & 83.54\% & 80.69\% \\
 & Zero-shot LLM & \color{low_score}51.78\% & 54.40\% & 57.36\% & 57.11\% & 57.01\% & 58.19\% & \color{low_score}56.82\% \\
 & RAG for LLM & 72.76\% & 73.60\% & 73.88\% & 74.31\% & 75.02\% & 75.73\% & 74.51\% \\
 & SMART & 86.74\% & 83.10\% & 85.67\% & 83.72\% & 84.76\% & 86.31\% & \color{high_score}84.71\% \\
\cmidrule{1-9}
\multirow{4}{*}{EVM} 
 & Fine-tuned Llama & 74.20\% & 66.28\% & 68.38\% & 70.72\% & 68.43\% & 74.73\% & 69.71\% \\
 & Zero-shot LLM & \color{low_score}58.05\% & 57.47\% & 59.93\% & 60.14\% & 60.14\% & 59.60\% & \color{low_score}59.46\% \\
 & RAG for LLM & 70.38\% & 68.47\% & 70.40\% & 70.41\% & 68.14\% & 70.98\% & 69.68\% \\
 & SMART & \color{high_score}76.79\% & 75.85\% & 78.99\% & 73.68\% & 74.60\% & 78.30\% & \color{high_score}76.28\% \\
\bottomrule
\end{tabular}
}

\caption{Comparison of baseline model's performance across multiple languages on \textbf{MultiTEND} dataset based on the metrics. Notice that \textbf{AVG} is the average value of the corresponding metric across the 5 \textbf{non-English} languages}
\label{appx.tab:Performence Comparison of baselines on MultiTEND}

\end{table*}

\begin{table}[th!]
  \centering
  \begin{tabular}{p{2cm}p{2.5cm}p{1.3cm}p{1.3cm}p{1.3cm}p{1.3cm}p{1.3cm}p{1.3cm}}
    \toprule
    \rowcolor{row_color}
    \textbf{Dataset} & \textbf{Model} & \textbf{EM} & \textbf{QSM} & \textbf{QFC} & \textbf{EX} & \textbf{EFM} & \textbf{EVM} \\
    \midrule
     TEND & SMART & 23.82\% & 63.21\% & 75.60\% & 65.08\% & 87.21\% & 72.79\% \\
    \bottomrule
  \end{tabular}
  \caption{Performance of SMART on \textbf{TEND} dataset based on the metrics.}
  \label{appx.tab:Performance of SMART on TEND}
\end{table}

To clarify the challenges posed by multilingual Text-to-NoSQL tasks to existing models, we first fine-tuned the Llama-3.2-1B model for testing and found its accuracy to be very low. Through detailed experiments (as shown in Figure~\ref{appx.tab:Performence Comparison of baselines on MultiTEND}), we further analyzed the additional challenges multilingual Text-to-NoSQL tasks pose to existing models. Results show that the common errors in NoSQL query generation from different models are primarily caused or worsened by multilingual issues. Additionally, we compared the performance of SMART~\cite{lu2024natural}, a model specifically designed for Text-to-NoSQL, on a pure English dataset (TEND) and a mixed dataset of six languages (MultiTEND)(See Table~ref{appx.tab:Performence Comparison of baselines on MultiTEND} and Table~\ref{appx.tab:Performance of SMART on TEND}). Results show that despite being designed for NoSQL generation, the quality of NoSQL queries generated by SMART significantly drops in multilingual contexts. Based on the experimental results, we analyzed the factors leading to the decline in NoSQL generation quality in multilingual environments, with the findings as follows:
\begin{itemize}[noitemsep, topsep=0pt, partopsep=0pt, parsep=0pt]
    \item \textbf{Finding 1}: In the process of identifying entity mentions in NLQ and mapping them to corresponding database fields (i.e., schema linking), significant lexical differences across languages, especially in those with more complex lexical formation rules (such as Japanese hiragana and katakana, Russian Cyrillic characters, and the morphological variations in German and French), impose higher demands on the model's language understanding capabilities, resulting in a significant drop in mapping accuracy. 
    \item \textbf{Finding 2}: The support for nested documents and array structures in NoSQL requires models to have a stronger understanding of database schemas in multilingual environments to handle complex nested fields or situations requiring array expansion. This makes the already challenging schema linking task even more difficult due to the multilingual context.
    \item \textbf{Finding 3}: In multilingual contexts, NLQs often exhibit vastly different syntactic structures due to language differences, significantly increasing the difficulty for models to comprehend multilingual questions. This also leads to errors in two critical tasks: mapping intentions to operators (intention mapping task) and mapping intentions to database fields (schema linking task).
\end{itemize}
Based on these experiments and findings, we categorize the challenges in MultiTEND into Structural Challenge and Lexical Challenge.

\section{More Implementation Details}
\label{More_Experimental_Details}
\paragraph{Dataset Construction} During the dataset construction process, we utilized the ``gpt-4o-mini-2024-07-18'' model to extend the dataset from English to multiple languages, with the parameter setting temperature = 0.0. 
\paragraph{MultiLink} When building the RAG vector library, we construct a corresponding vector library for the training set of each language. In the vector library construction process, the text-to-embedding model used is ``text-embedding-ada-002'', and the Faiss library is employed to build indexes for efficient vector search.

In the data augmentation phase of MultiLink, we use the ``DeepSeek-V3'' model to expand multilingual data pairs, with the parameter setting `temperature = 0.0'.

In the Parallel Multilingual Sketch-Schema Prediction phase of MultiLink, all SLMs are fine-tuned based on ``Llama-3.2-1B'' using a full-parameter fine-tuning strategy with AdamW as the optimizer. We set the hyperparameters for fine-tuning as follows: batch size = 2, learning rate = 5e-5, and epochs = 3. Additionally, we configure the gradient accumulation steps to 8 and set the maximum input token length to 2048. For the generation of SLMs, we use top-p = 0.7 and temperature = 0.0, with a maximum input token length of 2048 and a maximum output token length of 512.

In the Retrieval-Augmented Chain-of-Thought Query Generation phase of MultiLink, when retrieving examples from the vector library, we set the similarity threshold to 0.5 and the retrieval count (rag retrieve num) to 6.0. For query generation, the LLM used is ``DeepSeek-V3'', with the parameter setting `temperature = 0.0'.

\paragraph{Baselines} For baseline methods of Zero-shot LLM, Few-shot LLM, RAG for LLM, the LLM we use is ``DeepSeek-V3'' , with the parameter setting `temperature = 0.0'; Specifically for RAG for LLM, we employ the same vector library as MultiLink, setting the similarity threshold to 0.5 and the retrieval count (rag retrieve num) to 6.0.

For the part of SMART that requires the use of LLM, we also use ``DeepSeek-V3'', with the parameter setting` temperature = 0.0', which is consistent with the configuration of MultiLink. For the part of SMART that involves fine-tuning SLM as well as the baseline method of Fine-tuned SLM, we maintain the same settings as those used for fine-tuning SLM in MultiLink, using ``Llama-3.2-1B''. The hyperparameters for fine-tuning are set as follows: batch size = 2, learning rate = 5e-5, and epochs = 3. Additionally, we configure the gradient accumulation steps to 8 and set the maximum input token length to 2048. For the generation of SLMs, we use top-p = 0.7 and temperature = 0.0, with a maximum input token length of 2048 and a maximum output token length of 512.

\newpage
\section{More Experimental Details}
\label{More_Experimental_Details}

\subsection{Baselines}
\label{baseline}
We utilized a variety of popular neural network models, LLM-based prompting methods,SLM-based fine-tuning methods and existing Text-to-NoSQL pipelines as baseline models for a comprehensive performance comparison with MultiLink. The baseline models are as follows:

\begin{itemize}[noitemsep, topsep=0pt, partopsep=0pt, parsep=0pt]
    \item \textbf{Zero-shot LLM}: The zero-shot prompting approach utilizes the inherent zero-shot learning capabilities of LLM, allowing LLM to produce precise and contextually appropriate responses without the need for prior training or example-driven instructions.
    \item \textbf{Few-shot LLM}: The few-shot prompting technique serves as a key mechanism for in-context learning (ICL), where a limited set of examples is incorporated into the context to instruct LLM on executing tasks within specialized domains.
    \item \textbf{RAG for LLM}: Retrieval-Augmented Generation (RAG) technology provides an alternative approach to support LLM in downstream tasks. Unlike direct few-shot prompting, RAG dynamically retrieves relevant examples from a knowledge base based on the model's input, enriching the context and effectively reducing hallucinations induced by insufficient or ambiguous information.
    \item \textbf{Fine-tuned SLM}: Fine-tuning is another effective strategy for enhancing the performance of language models in specific downstream tasks, such as predicting NoSQL query generation. We fine tune SLM based on two different approaches (Monolingual Training and Multilingual Training) to compare the quality of NoSQL queries predicted by SLM based on single-target language training data and training data from multiple languages.
    \item \textbf{SMART}: SMART is the first and currently the only framework in the Text-to-NoSQL domain tackling the task of converting English NLQs to NoSQL queries. With the assistance of SLM and RAG technologies, it constructs four main processes: SLM-based schema prediction, SLM-based query generation, query refinement based on predicted schema and retrieved examples, and execution results-based query optimization.
    \item \textbf{MultiLink}: MultiLink is the framework proposed in this work, aiming to address the challenges of multilingual Text-to-NoSQL tasks. It constructs three main processes: Intention-aware Multilingual Data Augmentation (MIND),  Parallel Multilingual
Sketch-Schema Prediction,and Retrieval-Augmented Chain-of-Thought Query Prediction.
\end{itemize}

\subsection{Evaluation Metrics}
\label{metrics}
We report results using the same metrics as SMART, which include Exact Match (EM) and Execution Accuracy (EX), each with more detailed subdivisions such as Query Stages Match (QSM) and Query Fields Coverage (QFC) under EM, and Execution Fields Match (EFM) and Execution Value Match (EVM) under EX. 
\par Here are detailed descriptions of each metric:
\begin{itemize}[noitemsep, topsep=0pt, partopsep=0pt, parsep=0pt]
    \item \textbf{Exact Match (EM)}: The purpose of this metric is to evaluates whether the generated query is an exact match to the gold query, considering both its structure and content. It is calculated as:
\[ EM = \frac{N_{em}}{N} \]
    Where $N_{em}$ represents the count of queries fully matching the gold query, and N signifies the total number of queries within the test set. EM serves as a stringent measure of syntactic and semantic alignment.
    \begin{itemize}
        \item \textbf{Query Stages Match (QSM)}: QSM is designed to check if the generated query's key stages (e.g., match,group, lookup) mirror the gold query in the order and keywords employed. It's calculated as:
        \[ QSM = \frac{N_{qsm}}{N} \]
        where $N_{qsm}$ represents the count of queries with matching stages.
        \item \textbf{Query Fields Coverage (QFC)}: QFC assesses if the generated query encompasses all the fields present in the gold query, considering both database fields and  query-defined fields. It's defined as:
        \[ QFC = \frac{N_{qfc}}{N} \]
        Where $N_{qfc}$ represents the number of queries with complete field coverage.
    \end{itemize}
    
    \item \textbf{Execution Accuracy (EX)}: This metric evaluates the accuracy of the execution results for the generated query on the database. It is calculated as follows:
    \[ EX= \frac{N_{ex}}{N} \]
    where $N_{ex}$ represents the number of queries whose execution results align with those of the gold query. EX serves as the most critical performance metric for evaluating Text-to-NoSQL models.
    \begin{itemize}
        \item \textbf{Execution Fields Match (EFM)}: EFM validates the alignment of field names derived from the execution of the generated query against those obtained from the gold query. It's defined as:
        \[ EFM= \frac{N_{efm}}{N} \]
        where $N_{efm}$ represents the number of queries with matching field names in the results.
        \item \textbf{Execution Value Match (EVM)}: EVM evaluates the correspondence between the values in the execution results of the generated query and those in the gold query. It is defined as: 
        \[ EVM= \frac{N_{evm}}{N} \]
        where $N_{evm}$ represents the number of queries with matching values in the results.
    \end{itemize}
    
\end{itemize}
\newpage
\section{More Experimental Results}
\subsection{Performance Comparison}
\label{appendix: Performance Comparison}

\begin{table*}[th!]
\centering
{
\begin{tabular}{llccccccc}
\toprule
\textbf{Metric} & \textbf{Model} & \textbf{EN} & \textbf{ZH} & \textbf{FR} & \textbf{DE} & \textbf{JA} & \textbf{RU} & \textbf{AVG (5 langs)} \\
\midrule
\multicolumn{9}{c}{\textbf{Query-based Metric Results}} \\
\midrule
\multirow{6}{*}{EM} 
 & Fine-tuned Llama & 17.05\% & 13.57\% & 16.53\% & 15.78\% & 16.40\% & 14.51\% & 15.36\% \\
 & Zero-shot LLM & \color{low_score}0.29\% & \color{low_score}0.61\% & \color{low_score}0.61\% & \color{low_score}0.54\% & \color{low_score}0.54\% & \color{low_score}0.29\% & \color{low_score}0.52\% \\
 & Few-shot LLM & 12.18\% & 10.25\% & 10.65\% & 10.65\% & 9.87\% & 11.34\% & 10.55\% \\
 & RAG for LLM & 16.09\% & 13.98\% & 15.62\% & 14.33\% & 12.02\% & 13.89\% & 13.97\% \\
 & SMART & 18.85\% & 13.94\% & 18.38\% & 18.30\% & 18.05\% & 15.89\% & 16.91\% \\
 & MultiLink (Ours) & \color{high_score}30.05\% & \color{high_score}23.47\% & \color{high_score}25.58\% & \color{high_score}25.65\% & \color{high_score}23.53\% & \color{high_score}24.95\% & \color{high_score}24.64\% \\
\midrule
\multirow{6}{*}{QSM} 
 & Fine-tuned Llama & 57.19\% & 56.71\% & 54.22\% & 56.14\% & 56.22\% & 54.91\% & 55.64\% \\
 & Zero-shot LLM & \color{low_score}51.24\% & \color{low_score}47.76\% & \color{low_score}50.36\% & \color{low_score}50.43\% & \color{low_score}47.35\% & \color{low_score}48.95\% & \color{low_score}48.97\% \\
 & Few-shot LLM & 57.01\% & 58.52\% & 56.06\% & 53.83\% & 56.14\% & 55.45\% & 56.00\% \\
 & RAG for LLM & 62.30\% & 59.52\% & 60.51\% & 60.17\% & 57.36\% & 58.87\% & 59.28\% \\
 & SMART & 61.15\% & 57.69\% & 61.23\% & 59.35\% & 58.11\% & 57.17\% & 58.71\% \\
 & MultiLink (Ours) & \color{high_score}65.91\% & \color{high_score}62.19\% & \color{high_score}64.71\% & \color{high_score}64.73\% & \color{high_score}63.26\% & \color{high_score}63.29\% & \color{high_score}63.63\% \\
\midrule
\multirow{6}{*}{QFC} 
 & Fine-tuned Llama & 60.76\% & \color{low_score}53.83\% & \color{low_score}56.61\% & 62.35\% & 58.70\% & \color{low_score}58.41\% & \color{low_score}57.98\% \\
 & Zero-shot LLM & \color{low_score}60.29\% & 58.59\% & 58.92\% & 60.07\% & \color{low_score}58.09\% & 60.22\% & 59.18\% \\
 & Few-shot LLM & 62.88\% & 63.32\% & 59.39\% & \color{low_score}58.38\% & 61.51\% & 61.23\% & 60.76\% \\
 & RAG for LLM & 68.04\% & 67.86\% & 67.03\% & 67.47\% & 65.29\% & 67.46\% & 67.02\% \\
 & SMART & 65.05\% & 60.36\% & 60.97\% & 63.86\% & 62.03\% & 59.34\% & 61.31\% \\
 & MultiLink (Ours) & \color{high_score}76.55\% & \color{high_score}71.14\% & \color{high_score}73.22\% & \color{high_score}73.73\% & \color{high_score}72.18\% & \color{high_score}72.22\% & \color{high_score}72.50\% \\
\midrule
\multicolumn{9}{c}{\textbf{Execution-based Metric Results}} \\
\midrule
\multirow{6}{*}{EX} 
 & Fine-tuned Llama & 44.61\% & 36.86\% & 41.26\% & 41.44\% & 43.32\% & 38.23\% & 40.22\% \\
 & Zero-shot LLM & \color{low_score}36.58\% & \color{low_score}28.99\% & \color{low_score}33.86\% & \color{low_score}34.91\% & \color{low_score}30.63\% & \color{low_score}29.68\% & \color{low_score}31.61\% \\
 & Few-shot LLM & 40.79\% & 34.95\% & 36.64\% & 37.08\% & 35.93\% & 34.77\% & 35.87\% \\
 & RAG for LLM & 51.70\% & 47.02\% & 49.28\% & 48.59\% & 45.12\% & 45.99\% & 47.20\% \\
 & SMART & 48.86\% & 38.05\% & 44.69\% & 44.22\% & 43.30\% & 41.03\% & 42.26\% \\
 & MultiLink (Ours) & \color{high_score}67.64\% & \color{high_score}57.71\% & \color{high_score}59.86\% & \color{high_score}58.90\% & \color{high_score}55.75\% & \color{high_score}54.88\% & \color{high_score}57.42\% \\
\midrule
\multirow{6}{*}{EFM} 
 & Fine-tuned Llama & 84.97\% & 78.84\% & 80.14\% & 81.44\% & 79.50\% & 83.54\% & 80.69\% \\
 & Zero-shot LLM & \color{low_score}51.78\% & \color{low_score}54.40\% & \color{low_score}57.36\% & \color{low_score}57.11\% & \color{low_score}57.01\% & \color{low_score}58.19\% & \color{low_score}56.82\% \\
 & Few-shot LLM & 63.21\% & 63.79\% & 62.85\% & 65.31\% & 64.47\% & 66.97\% & 64.68\% \\
 & RAG for LLM & 72.76\% & 73.60\% & 73.88\% & 74.31\% & 75.02\% & 75.73\% & 74.51\% \\
 & SMART & 86.74\% & 83.10\% & \color{high_score}85.67\% & 83.72\% & 84.76\% & \color{high_score}86.31\% & 84.71\% \\
 & MultiLink (Ours) & \color{high_score}88.92\% & \color{high_score}85.41\% & 84.64\% & \color{high_score}85.27\% & \color{high_score}85.32\% & 84.40\% & \color{high_score}85.01\% \\
\midrule
\multirow{6}{*}{EVM} 
 & Fine-tuned Llama & 74.20\% & 66.28\% & 68.38\% & 70.72\% & 68.43\% & 74.73\% & 69.71\% \\
 & Zero-shot LLM & \color{low_score}58.05\% & \color{low_score}57.47\% & \color{low_score}59.93\% & \color{low_score}60.14\% & \color{low_score}60.14\% & \color{low_score}59.60\% & \color{low_score}59.46\% \\
 & Few-shot LLM & 65.37\% & 63.21\% & 63.94\% & 66.35\% & 63.46\% & 65.96\% & 64.58\% \\
 & RAG for LLM & 70.38\% & 68.47\% & 70.40\% & 70.41\% & 68.14\% & 70.98\% & 69.68\% \\
 & SMART & \color{high_score}76.79\% & \color{high_score}75.85\% & \color{high_score}78.99\% & \color{high_score}73.68\% & \color{high_score}74.60\% & \color{high_score}78.30\% & \color{high_score}76.28\% \\
 & MultiLink (Ours) & 76.33\% & 74.32\% & 74.42\% & 73.55\% & 72.03\% & 73.38\% & 73.54\% \\
\bottomrule
\end{tabular}
}
\caption{Comparison of each model's performance across multiple languages on MultiTEND based on the metrics. Notice that \textbf{AVG} is the average value of the corresponding metric across the 5 \textbf{non-English} languages.}
\label{appx.tab:Performence Comparison}
\end{table*} 

\begin{table}[th!]
  \centering
  \begin{tabular}{lcccccc}
    \toprule
    Model & EM & QSM & QFC & EX & EFM & EVM \\
    \midrule
    Fine-tuned Llama & 15.64\% & 55.90\% & \color{low_score}58.44\% & 40.95\% & 81.41\% & 70.46\% \\
    Zero-shot LLM & \color{low_score}0.48\% & \color{low_score}49.35\% & 59.36\% & \color{low_score}32.44\% & \color{low_score}55.98\% & \color{low_score}59.22\% \\
    Few-shot LLM & 10.82\% & 56.17\% & 61.12\% & 36.69\% & 64.43\% & 64.71\% \\
    RAG for LLM & 14.32\% & 59.79\% & 67.19\% & 47.95\% & 74.22\% & 69.80\% \\
    SMART & 17.23\% & 59.12\% & 61.94\% & 43.36\% & 85.05\% & \color{high_score}\textbf{76.37\%} \\
    \textbf{MultiLink (Ours)} & \color{high_score}\textbf{25.54\%} & \color{high_score}\textbf{64.01\%} & \color{high_score}\textbf{73.17\%} & \color{high_score}\textbf{59.12\%} & \color{high_score}\textbf{85.66\%} & 74.01\% \\
    \bottomrule
  \end{tabular}
  \caption{Comparison of each model's average performance across six languages(\textbf{AVG of 6 langs}) on MultiTEND based on the metrics.}
  \label{appx.tab:Performance Comparison(6 langs)}
\end{table}

As shown in Table \ref{appx.tab:Performence Comparison}, models consistently achieve higher performance in English across all metrics, suggesting baseline models are more proficient in handling complex tasks in English. This can be attributed to their extensive training on English data, giving them an inherent advantage over other languages. In contrast, performance in Chinese and Russian often lags behind languages like English, Japanese, French, and German. This disparity may arise not only from limited training data for these languages but also from the unique challenges posed by Chinese character construction, syntax, and the Cyrillic alphabet in Russian, which introduce additional complexities in comprehension and generation.

According to the results shown in Table \ref{appx.tab:Performence Comparison} and Table \ref{appx.tab:Performance Comparison(6 langs)}, there are significant differences in the performance of various models on key metrics. Our model performs exceptionally well across all languages and metrics except for EVM, outperforming the second-best baseline model by approximately 5\% to 20\%. Particularly on the two key metrics for real-world applications, EM and EX, our model maintains an absolute lead with average advantages of 25.54\% and 59.12\%, respectively, across all languages.
In contrast, the Zero-shot LLM method performs the worst across all languages and metrics, particularly on EM and EX, with its average accuracy across six languages trailing MultiLink by nearly 25\%. The other three methods achieve average accuracies across the six languages on EM of 15.64\% (Fine-tuned), 14.32\% (RAG for LLM), and 10.82\% (Zero-shot LLM), while on EX, their average accuracies are 47.95\% (RAG for LLM), 40.95\% (Fine-tuned Llama), and 36.69\% (Few-shot LLM). Overall, our model demonstrates significant advantages across all metrics on every language.

Further analysis of Table \ref{appx.tab:Performance Comparison(6 langs)} reveals distinct performance differences between model fine-tuning and direct prompting LLM methods on the MultiTEND test set under cross-domain criteria. Fine-tuned Llama achieved an 81.41\% accuracy in EFM, outperforming RAG for LLM by 7.19\%, indicating its stronger capability in generating queries that retrieve correct field results. On the other hand, RAG for LLM surpassed Fine-tuned Llama by margins ranging from 3.8\% to 8.75\% in EX, QSM, and QFC metrics, demonstrating its superior understanding of the mapping relationships between NLQs and NoSQL database fields, as well as a deeper grasp of data operations in NoSQL queries, leading to higher execution accuracy

Among the Multi-Step methods specifically designed for Text-to-NoSQL tasks, the performance gap between SMART and our proposed MultiLink approach is significant. As shown in Table \ref{appx.tab:Performence Comparison}, specifically, in terms of performance across every language, MultiLink outperforms SMART, which is tailored for monolingual NoSQL generation, across all metrics except EVM, with margins ranging from 2.18\% to 18.78\%. This demonstrates that SMART, designed exclusively for English contexts, cannot directly address the challenges posed by multilingual tasks. In contrast, the intention mapping and schema linking methods designed in MultiLink, specifically developed to tackle multilingual generation challenges, effectively overcome these difficulties, enabling it to achieve significantly better performance in multilingual scenarios compared to other baseline models.

\newpage
\subsection{Parameter Study}
\label{appendix: Parameter Study}
\begin{figure*}[ht]
    \centering
    \begin{subfigure}[b]{0.32\textwidth}
        \centering
        \includegraphics[width=\textwidth]{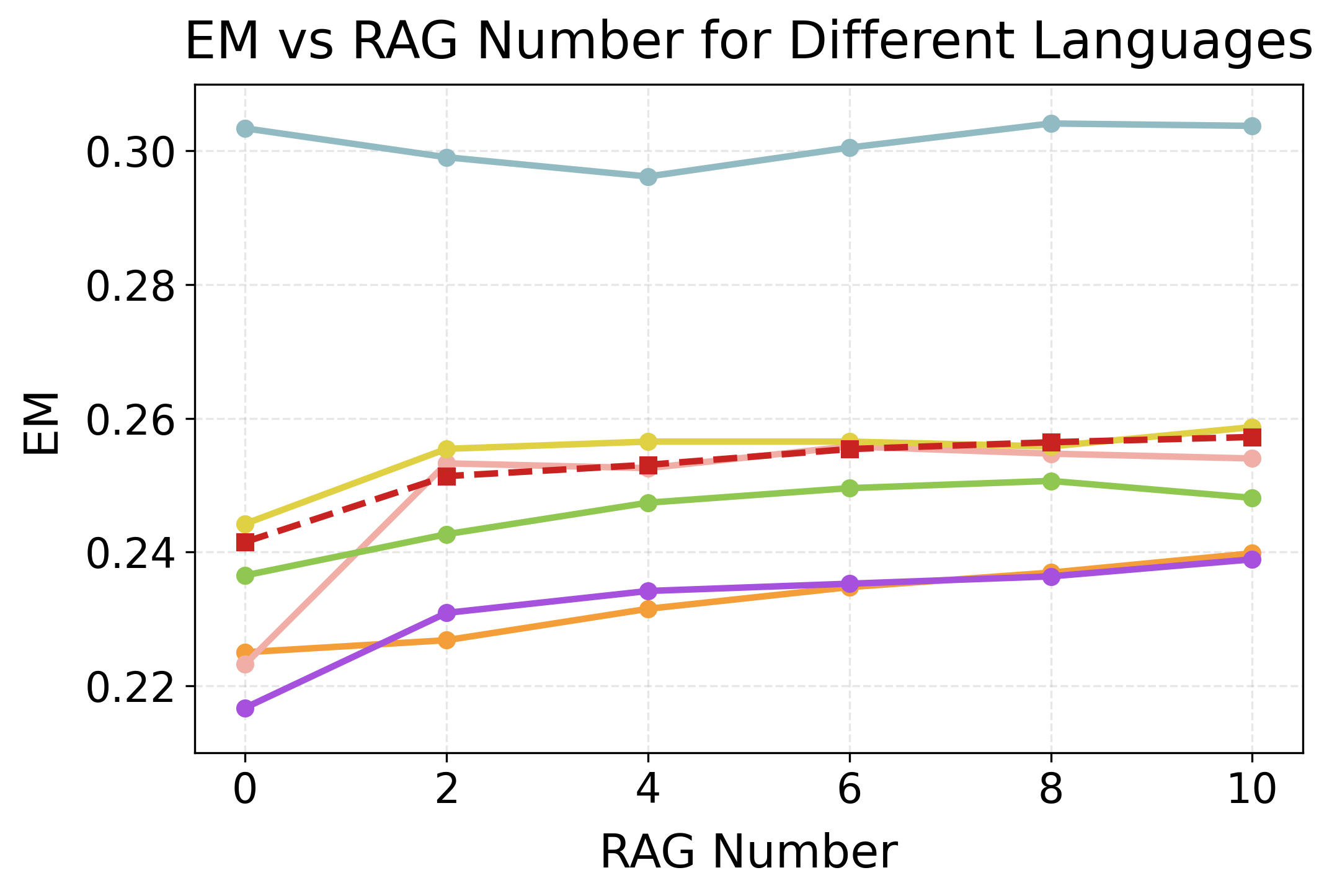}
        \caption{EM Score}
        \label{fig:em}
    \end{subfigure}
    \hfill
    \begin{subfigure}[b]{0.32\textwidth}
        \centering
        \includegraphics[width=\textwidth]{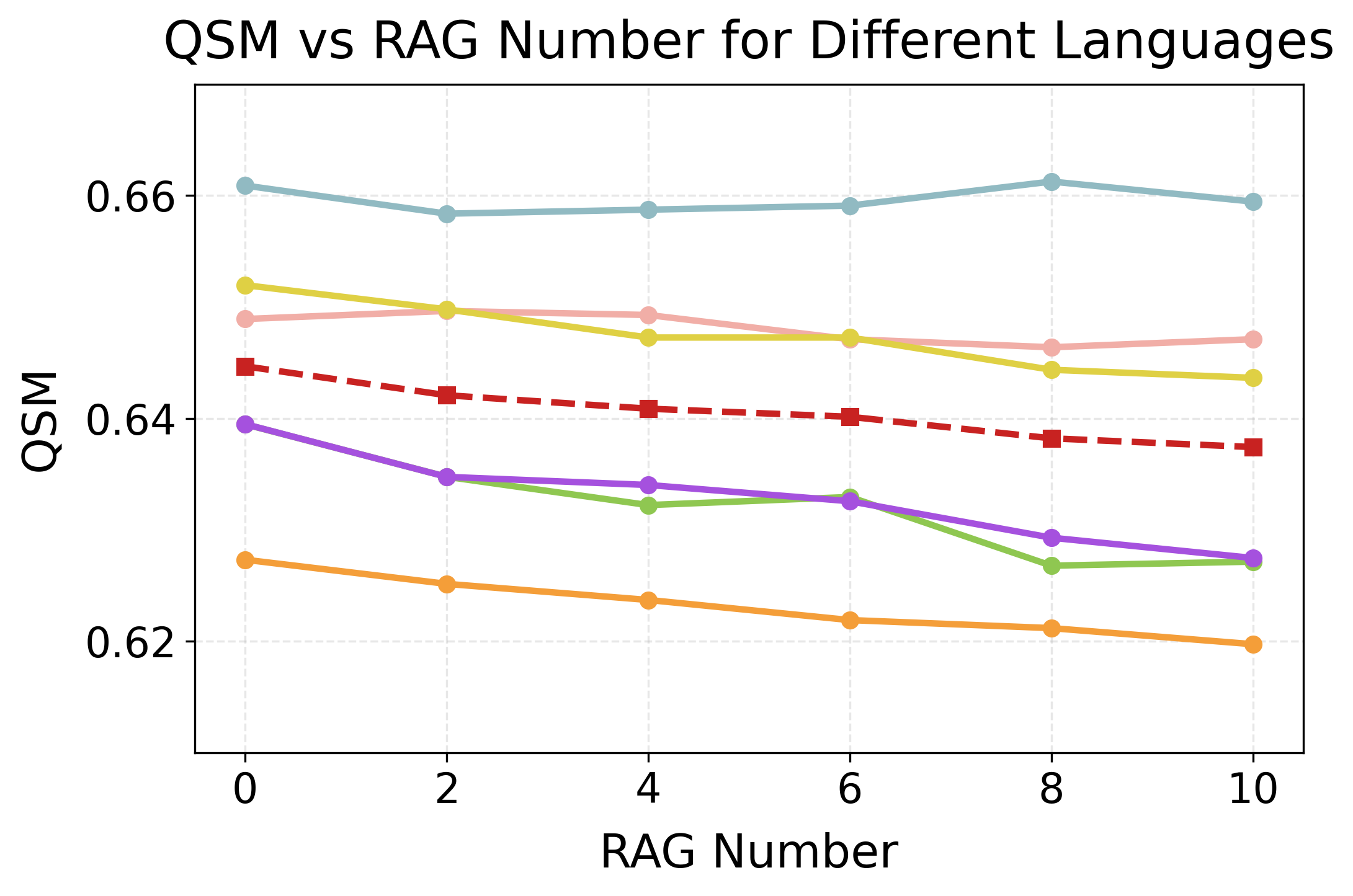}
        \caption{QSM Score}
        \label{fig:qsm}
    \end{subfigure}
    \hfill
    \begin{subfigure}[b]{0.32\textwidth}
        \centering
        \includegraphics[width=\textwidth]{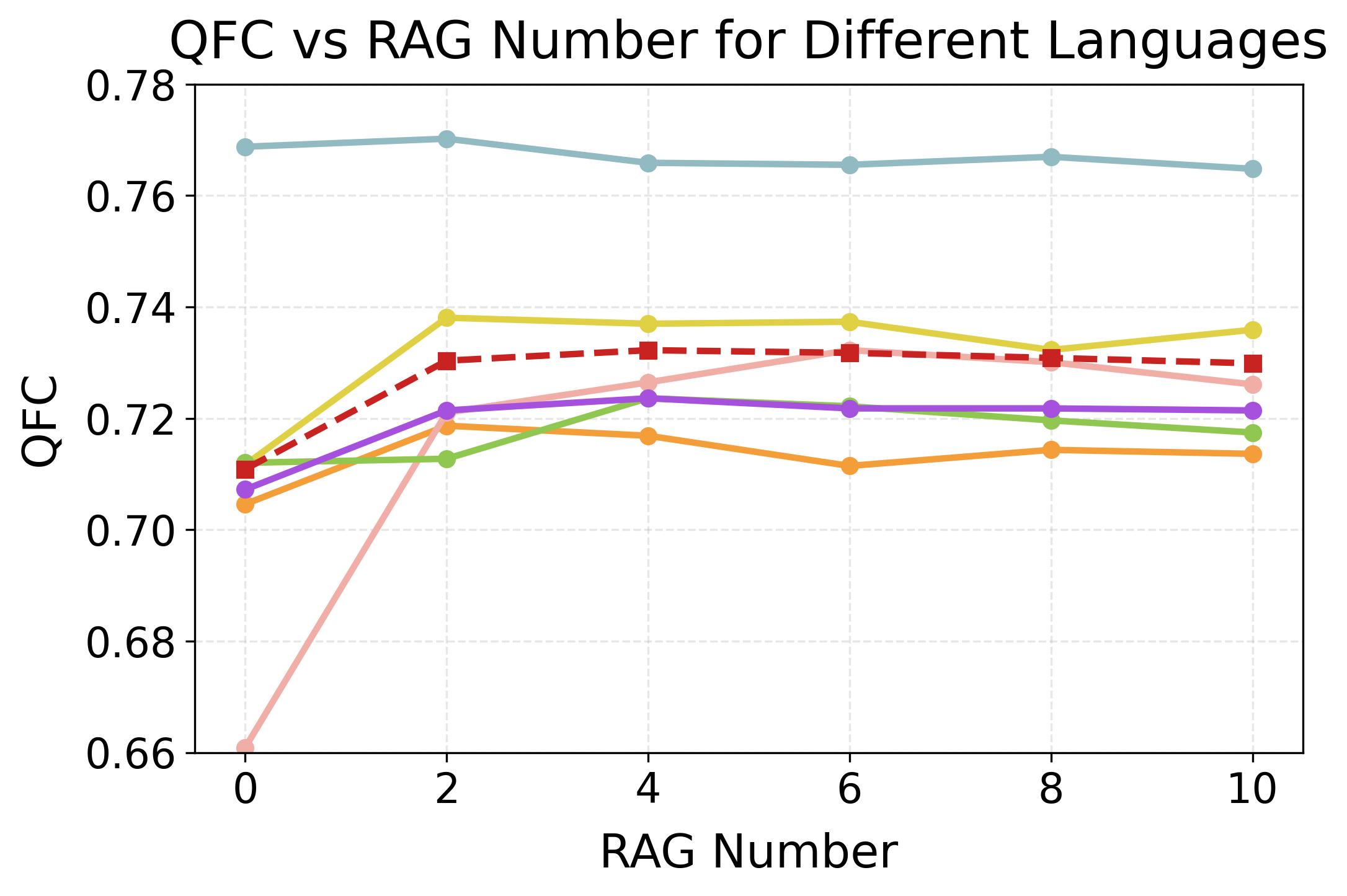}
        \caption{QFC Score}
        \label{fig:qfc}
    \end{subfigure}
    
    \vspace{0.2cm}
    
    \begin{subfigure}[b]{0.32\textwidth}
        \centering
        \includegraphics[width=\textwidth]{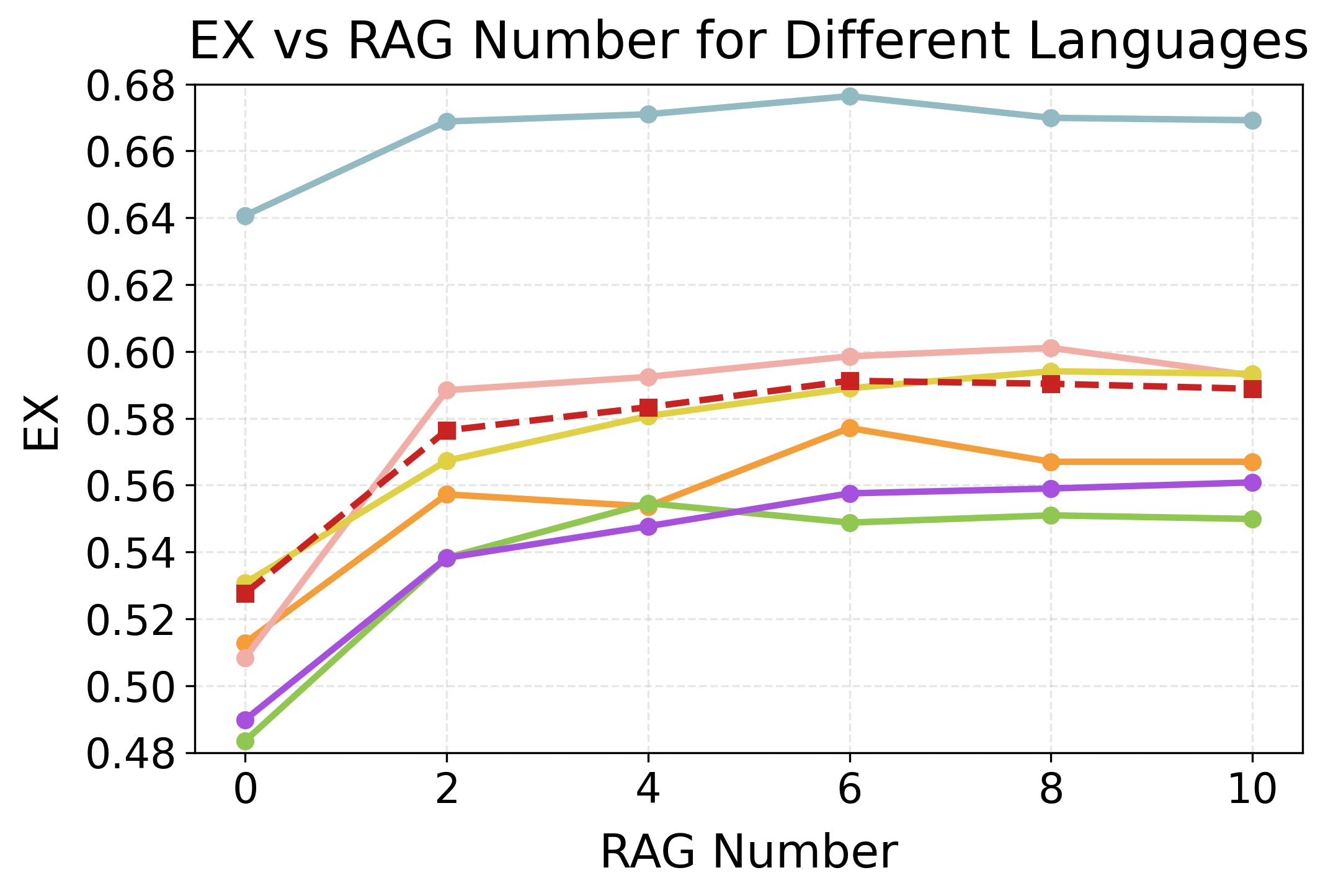}
        \caption{EX Score}
        \label{fig:ex}
    \end{subfigure}
    \hfill
    \begin{subfigure}[b]{0.32\textwidth}
        \centering
        \includegraphics[width=\textwidth]{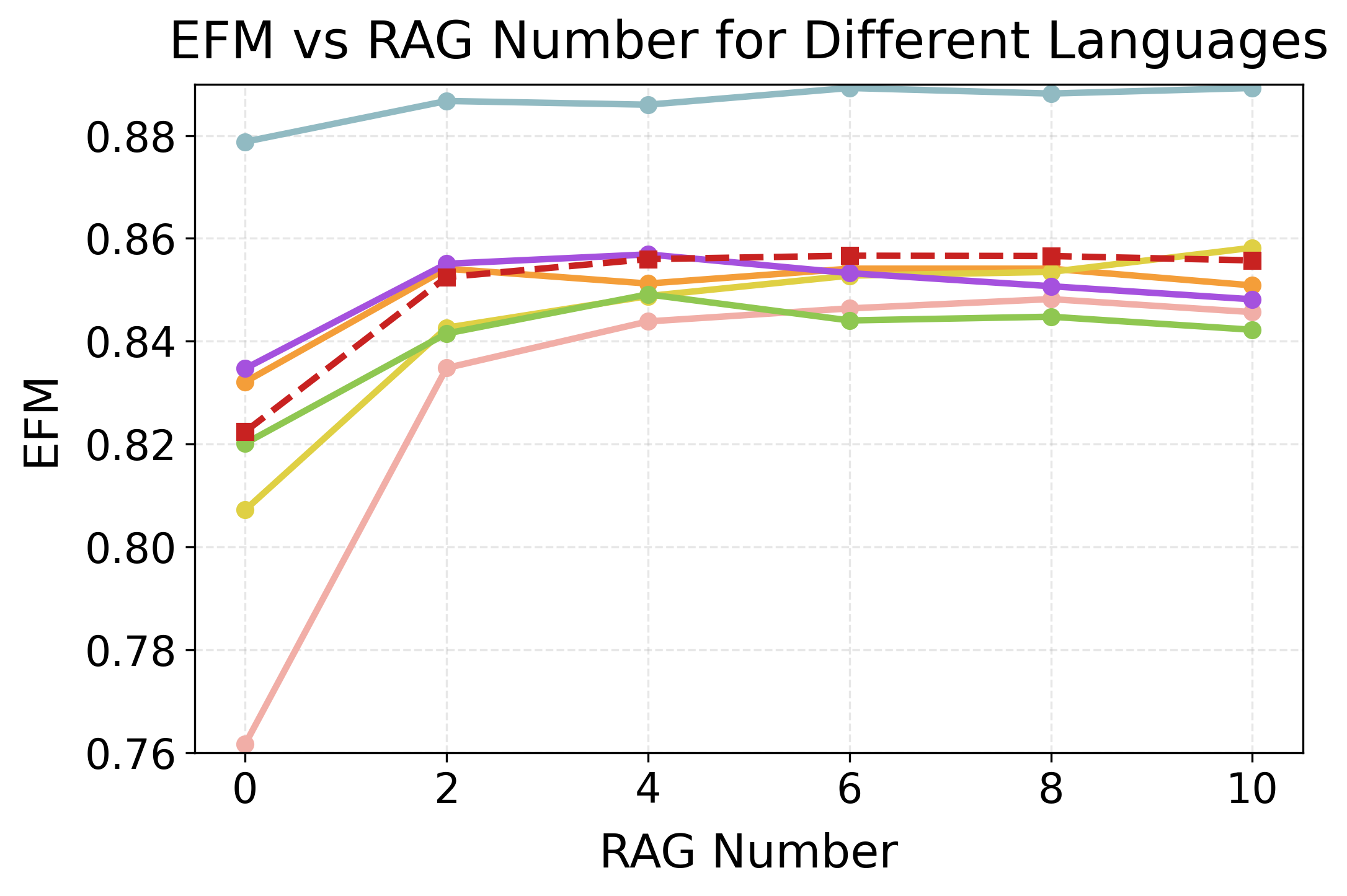}
        \caption{EFM Score}
        \label{fig:efm}
    \end{subfigure}
    \hfill
    \begin{subfigure}[b]{0.32\textwidth}
        \centering
        \includegraphics[width=\textwidth]{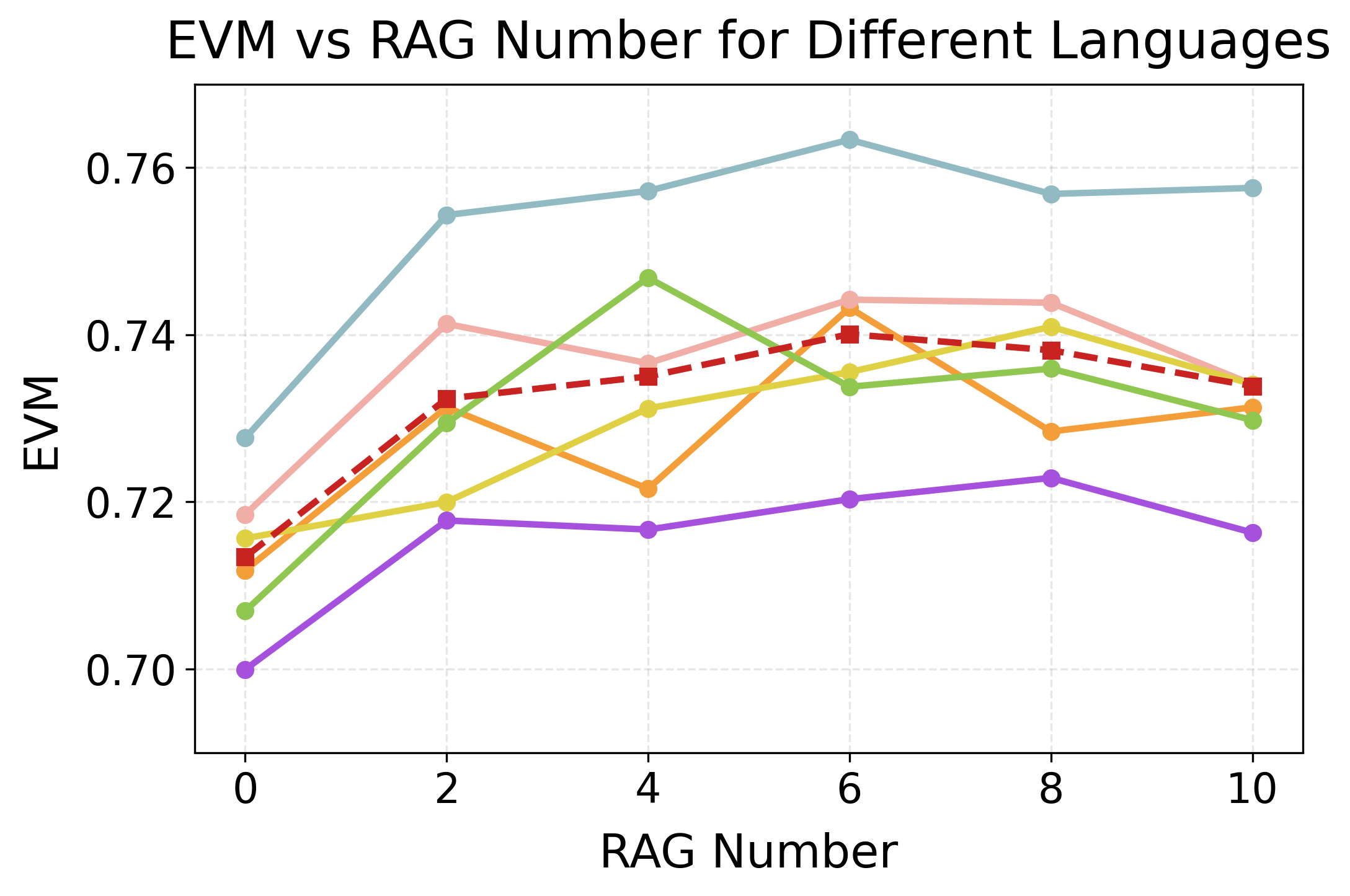}
        \caption{EVM Score}
        \label{fig:evm}
    \end{subfigure}
    
    \includegraphics[width=0.9\textwidth]{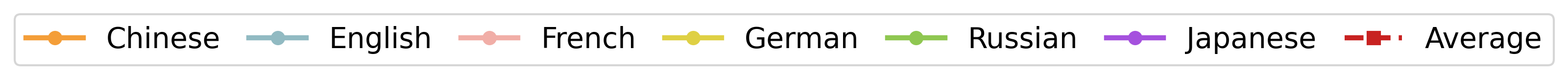}
    \vspace{-0.2cm}
    \caption{Parameter study of RAG number on the MultiTEND test set.}
    \label{appx.fig:parameter_study}
\end{figure*}
Figure \ref{appx.fig:parameter_study} illustrates the performance variations of MultiLink across multiple languages under different numbers of retrieved examples for all metrics. The vertical axis represents the model's accuracy under each metric, and the horizontal axis represents the RAG number ranging from 0 to 10. 

Analyzing Figure \ref{appx.fig:parameter_study} and observing the changes in the average represented by the red dashed line, we find that as the RAG num increases, MultiLink shows significant improvement on all metrics except QSM, followed by a slow rise and a gradual decline after reaching a certain value. The initial improvement, resulting from the change in RAG num from 0 to 2, indicates that RAG greatly enhances model performance. The subsequent decline in MultiLink's performance beyond a certain value might be attributed to excessive context length introducing redundant information, which interferes with generation.

Moreover, the model's performance on the QSM metric declines as the RAG num increases, which may be attributed to the influence of retrieved examples on the model's decisions regarding NoSQL operations. However, considering the performance on other metrics, the retrieved examples contribute to the model generating more accurate queries before the RAG num reaches 6, with the peak accuracy of model execution occurring at a RAG num of 6.

\newpage
\begin{CJK*}{UTF8}{gbsn}
\subsection{Case Study}
\label{appendix: Case Study}
Table \ref{tab:case_study} presents a case study comparing various baseline methods and MultiLink in generating NoSQL queries and their execution results. Firstly, we observe that the fine-tuned Llama exhibits a significant lack of understanding regarding database field names and overall structure, as illustrated in Table \ref{tab:case_study}. The model not only fails to accurately match target fields (such as ``课程名称''), leading to erroneous query logic and invalid results, but also incorrectly identifies ``课程'' as an independent collection rather than recognizing it as a field nested within the ``科目'' collection. Additionally, it lacks the ability to deconstruct nested fields (e.g., using \$unwind), rendering it incapable of properly handling array-type nested structures.

On the other hand, RAG for LLM performs well in handling the syntax of complex operations, but it still falls short in understanding field names and matching database schemas, and it fails to accurately incorporate the actual values in the database for retrieval. As shown in Table \ref{tab:case_study}, although the model correctly uses \texttt{\$unwind} to deconstruct nested fields, the query still cannot execute correctly because it does not filter based on the actual values in the database.

In comparison, SMART demonstrates a solid grasp of NoSQL syntax and shows a slightly better understanding of user query intentions and database structures compared to methods that directly utilize LLMs through prompting for generation. However, it struggles with handling more complex fields that involve nested relationships. Due to a lack of multilingual knowledge, it often incorrectly maps NLQ to database fields, finding it difficult to distinguish between semantically similar fields in different languages. As illustrated in Table \ref{tab:case_study}, SMART erroneously selected the ``课程'' collection instead of the ``科目'' collection and failed to correctly structure the nested fields. In contrast, MultiLink fully and accurately comprehended the query intent even in multilingual contexts, with clear query logic and the ability to generate queries that precisely align with the query intent. This indicates that MultiLink can effectively understand and execute multilingual Text-to-NoSQL tasks and generalizes well to test set examples after learning from the training set.
\end{CJK*}

\begin{CJK*}{UTF8}{gbsn}
\begin{table*}[th!]
    \centering
    \definecolor{tabletext}{RGB}{15, 23, 42}      
    \definecolor{resultcolor}{RGB}{13, 91, 41}    
    \definecolor{errorcolor}{RGB}{220, 38, 38}     
    \definecolor{syntaxcolor}{RGB}{216, 85, 85}    
    \definecolor{keywordcolor}{RGB}{20, 43, 144}   
    \definecolor{codebg}{RGB}{244, 245, 245}        
    \definecolor{successbg}{RGB}{239, 246, 247}    
    \definecolor{commentcolor}{RGB}{51, 65, 85}     
    \definecolor{nlqbg}{RGB}{126, 181, 189}         
    \definecolor{headerbg}{RGB}{179, 215, 174}      
    \definecolor{NLQcolor}{RGB}{255, 255, 255}
    
    \renewcommand{\arraystretch}{1.0}  
    \setlength{\tabcolsep}{3pt}        
    \begin{adjustbox}{width=\textwidth}
    \begin{tabular}{>{\raggedright\arraybackslash}p{0.13\linewidth}!{\color{black}\vrule width 0.5pt}>{\raggedright\arraybackslash}p{0.60\linewidth}!{\color{black}\vrule width 0.5pt}>{\raggedright\arraybackslash}p{0.28\linewidth}}
        \toprule[1pt]
        \rowcolor{nlqbg}
        \multicolumn{3}{l}{\small\rmfamily\textcolor{NLQcolor}{\textbf{NLQ: 学生在西班牙语课程的注册日期是什么时候？ (When is the registration date for students in the Spanish course?)}}} \\ [-0.5pt]
        \rowcolor{nlqbg}\midrule[0.2pt]
        \rowcolor{successbg}
        \textbf{Target NoSQL} & 
        \ttfamily\footnotesize
        \begin{minipage}[t]{\dimexpr\linewidth-2\fboxsep\relax}
        \raggedright
        db.科目.\textcolor{keywordcolor}{aggregate}([
        \{ \textcolor{keywordcolor}{\$unwind}: ``\$课程'' \}, 
        \{ \textcolor{keywordcolor}{\$match}: \{ ``课程.课程名称'': ``Spanish'' \} \}, 
        \{ \textcolor{keywordcolor}{\$unwind}: ``\$课程.学生课程注册'' \}, 
        \{ \textcolor{keywordcolor}{\$project}: \{ 注册的日期: ``\$课程.学生课程注册.注册的日期'', \_id: 0 \} \} 
        ]);
        \end{minipage} &
        \textcolor{commentcolor}{\footnotesize Standard query: Correct usage of nested field paths} \newline
        {\color{resultcolor}Results: \footnotesize [\{`注册的日期' : `2017-12-07 02:21:13'\},\dots]} \\
        \midrule[0.2pt]
        \rowcolor{codebg}
        \textbf{Fine-tuned Llama} & 
        \ttfamily\footnotesize\colorbox{codebg}{%
        \begin{minipage}[t]{\dimexpr\linewidth-2\fboxsep\relax}
        \raggedright
        db.\textcolor{errorcolor}{课程}.\textcolor{keywordcolor}{aggregate}([
        \{ \textcolor{keywordcolor}{\$match}: \{ 课程名称: ``Spanish'' \} \}, 
        \{ \textcolor{keywordcolor}{\$unwind}: ``\$学生课程注册'' \}, 
        \{ \textcolor{keywordcolor}{\$project}: \{ 注册的日期: ``\$学生课程注册.注册的日期'', \_id: 0 \} \} 
        ]);
        \end{minipage}} &
        \textcolor{commentcolor}{\footnotesize Error: Incorrect collection (`科目' vs. `课程')} \newline
        {\color{errorcolor}Results: \footnotesize []} \\
        \midrule[0.2pt]
        \rowcolor{codebg}
        \textbf{RAG for LLM} & 
        \ttfamily\footnotesize\colorbox{codebg}{%
        \begin{minipage}[t]{\dimexpr\linewidth-2\fboxsep\relax}
        \raggedright
        db.科目.\textcolor{keywordcolor}{aggregate}([
        \{ \textcolor{keywordcolor}{\$unwind}: ``\$课程'' \}, 
        \{ \textcolor{keywordcolor}{\$match}: \{ ``课程.课程名称'': \textcolor{errorcolor}{``西班牙语''} \} \}, 
        \{ \textcolor{keywordcolor}{\$unwind}: ``\$课程.学生课程注册'' \}, 
        \{ \textcolor{keywordcolor}{\$project}: \{ 注册日期: ``\$课程.学生课程注册.注册的日期'', \_id: 0 \} \} 
        ]);
        \end{minipage}} &
        \textcolor{commentcolor}{\footnotesize Error: Wrong value (`Spanish' vs. `西班牙语')} \newline
        {\color{errorcolor}Results: \footnotesize []} \\
        \midrule[0.2pt]
        \rowcolor{codebg}
        \textbf{SMART} & 
        \ttfamily\footnotesize\colorbox{codebg}{%
        \begin{minipage}[t]{\dimexpr\linewidth-2\fboxsep\relax}
        \raggedright
        db.\textcolor{errorcolor}{课程}.\textcolor{keywordcolor}{aggregate}([
        \{ \textcolor{keywordcolor}{\$match}: \{ 课程名称: ``Spanish'' \} \}, 
        \{ \textcolor{keywordcolor}{\$unwind}: ``\$学生课程注册'' \}, 
        \{ \textcolor{keywordcolor}{\$project}: \{ 注册的日期: ``\$学生课程注册.注册的日期'', \_id: 0 \} \} 
        ]);
        \end{minipage}} &
        \textcolor{commentcolor}{\footnotesize Error: Wrong collection and field paths} \newline
        {\color{errorcolor}Results: \footnotesize []} \\
        \midrule[0.2pt]
        \rowcolor{successbg}
        \textbf{MultiLink (Ours)} & 
        \ttfamily\footnotesize
        \begin{minipage}[t]{\dimexpr\linewidth-2\fboxsep\relax}
        \raggedright
        db.科目.\textcolor{keywordcolor}{aggregate}([
        \{ \textcolor{keywordcolor}{\$unwind}: ``\$课程'' \}, 
        \{ \textcolor{keywordcolor}{\$match}: \{ ``课程.课程名称'': ``Spanish'' \} \}, 
        \{ \textcolor{keywordcolor}{\$unwind}: ``\$课程.学生课程注册'' \}, 
        \{ \textcolor{keywordcolor}{\$project}: \{ 注册的日期: ``\$课程.学生课程注册.注册的日期'', \_id: 0 \} \} 
        ]);
        \end{minipage} &
        \textcolor{commentcolor}{\footnotesize Success: Correct structure and values} \newline
        {\color{resultcolor}Results: \footnotesize [\{`注册的日期' : `2017-12-07 02:21:13'\},\dots]} \\
        \bottomrule[1pt]
    \end{tabular}
    \end{adjustbox}
    \caption{\textbf{Case Study: Comparison of Different Approaches in Complex Nested Queries.} This table illustrates the performance of three baseline methods against our MultiLink method in generating Chinese MongoDB queries based on the same NLQ.}
    \label{tab:case_study}
\end{table*}
\end{CJK*}

\newpage
\section{Prompt Examples}
In this section, we present the specific prompts designed for each LLM application scenario within MultiLink.
\subsection{Prompt Design in Data Construction Pipeline}
\label{appendix:prompt in dataset}
\subsubsection{DB Fields Translation in Dataset Construction}
\begin{figure}[h!]
    \centering  
    \includegraphics[width=1.0\textwidth]{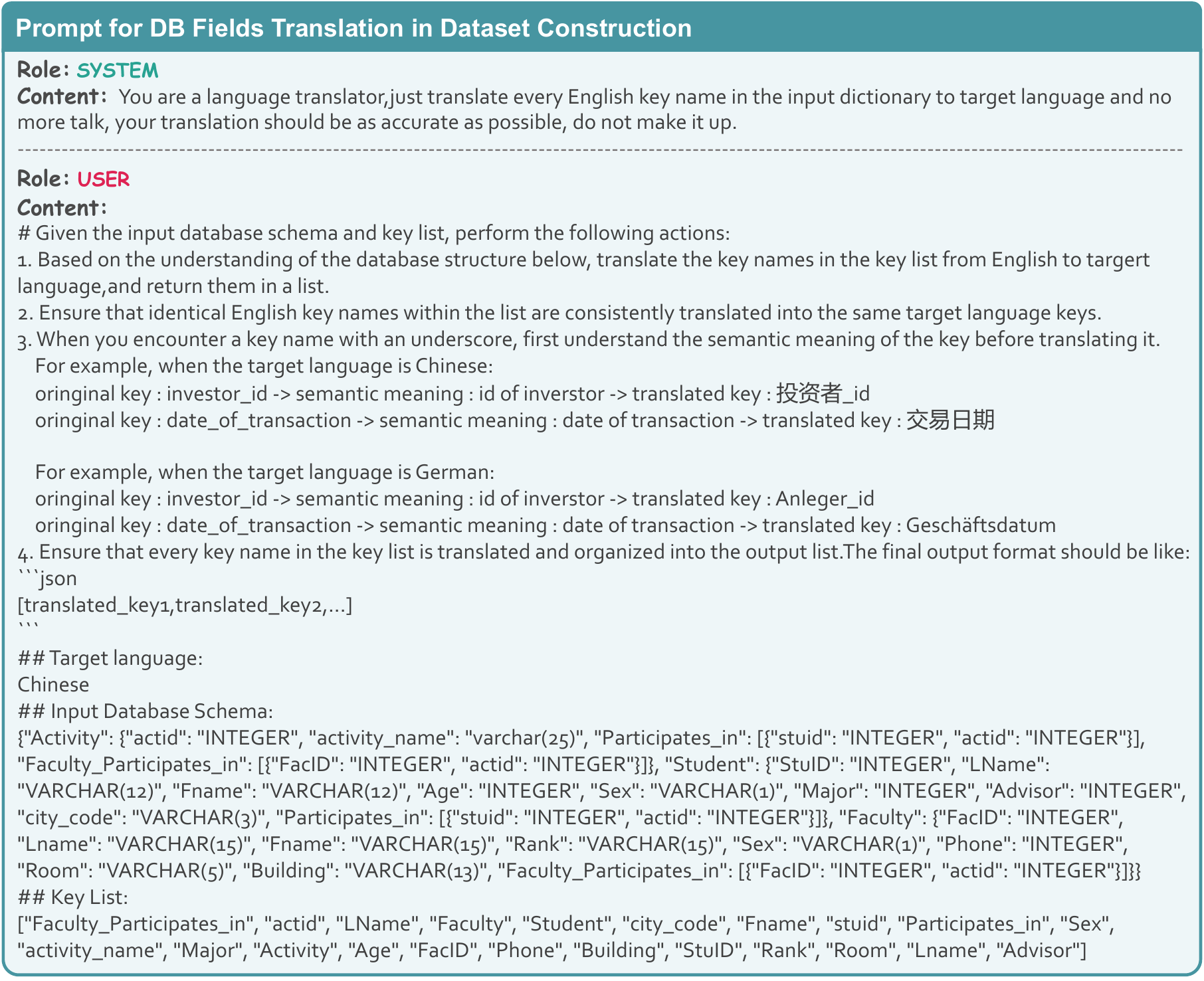}
\end{figure}

\newpage
\subsubsection{NLQ Translation in Dataset Construction}
\begin{figure*}[th!]
    \centering  
    \includegraphics[width=1.0\textwidth]{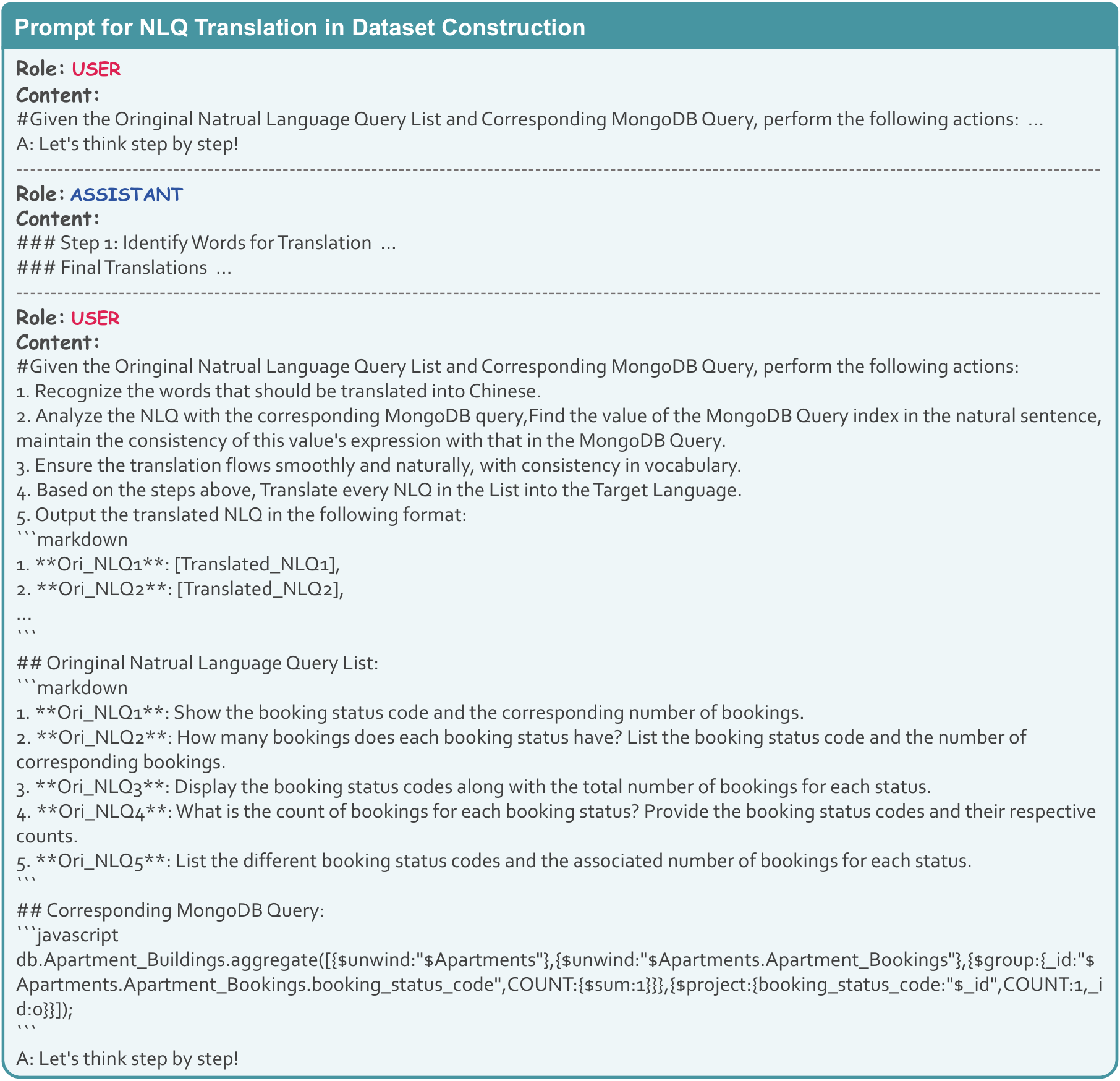}
\end{figure*}

\newpage
\subsection{Prompt Design in MultiLink}
\label{appendix:prompt in MultiLink}
\subsubsection{Intention-aware Multilingual Data Augmentation (MIND)}
\begin{figure*}[th!]
    \centering  
    \includegraphics[width=1.0\textwidth]{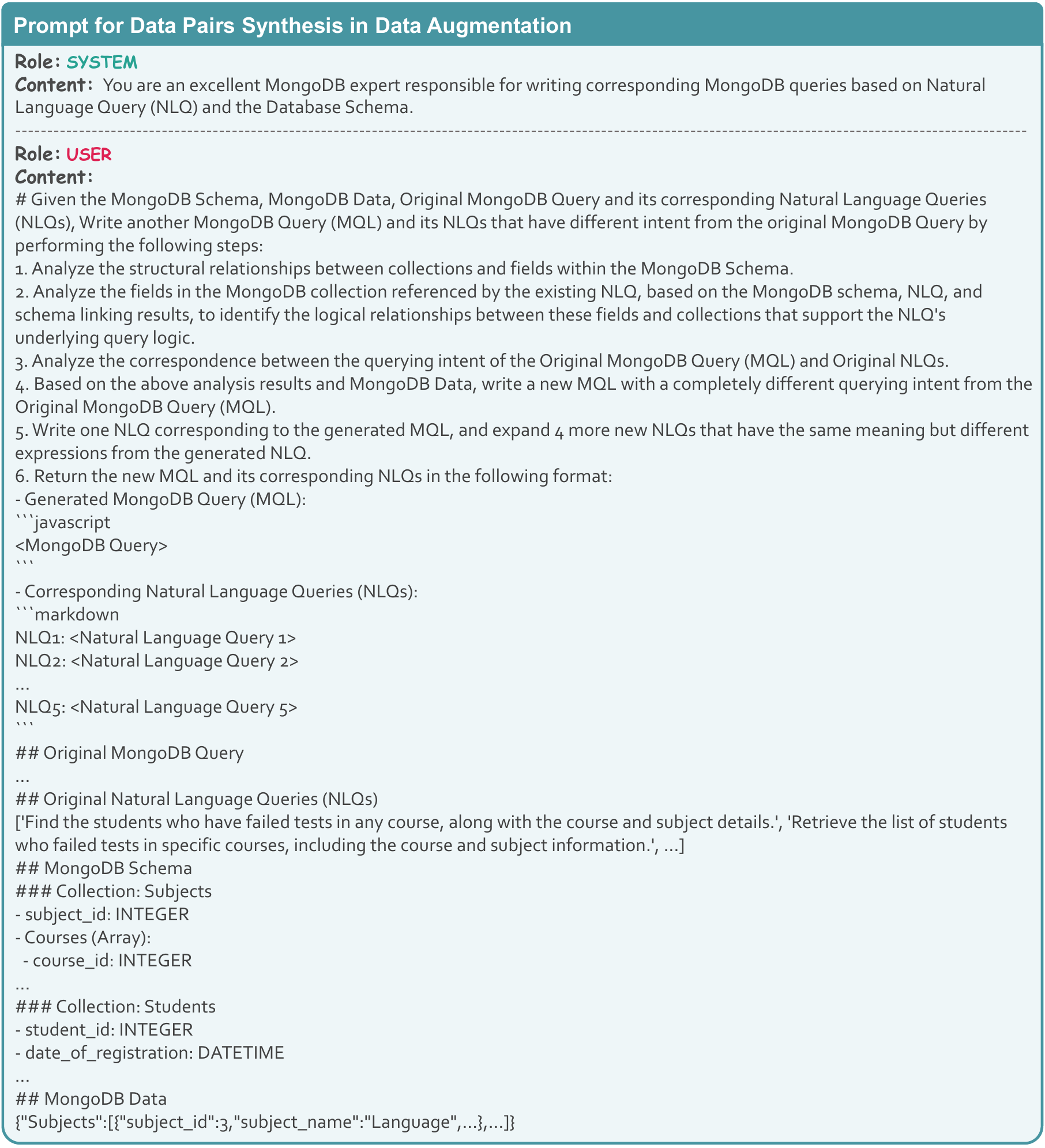}
\end{figure*}
\newpage
\begin{figure*}[th!]
    \centering  
    \includegraphics[width=1.0\textwidth]{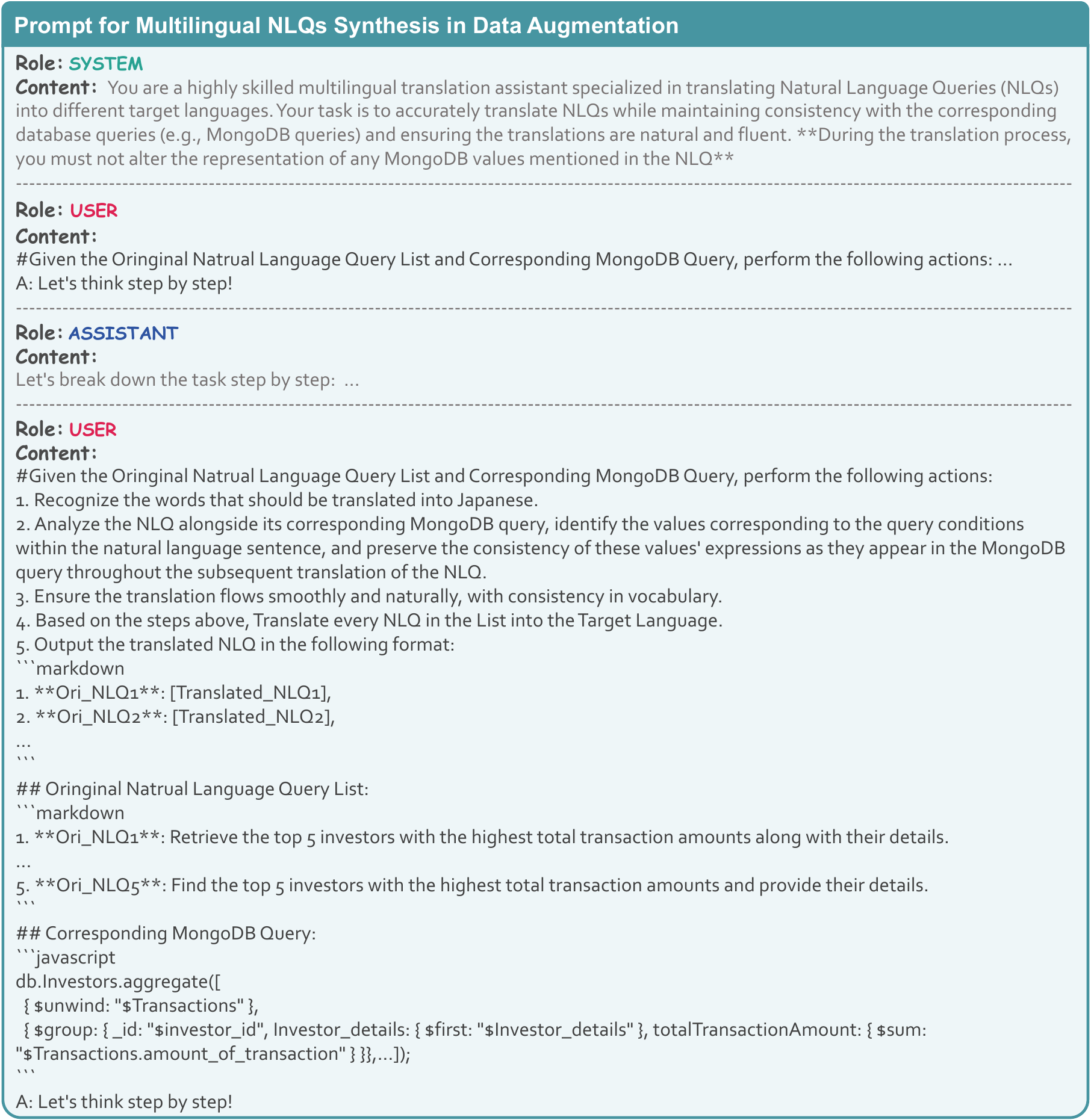}
\end{figure*}

\newpage
\subsubsection{Retrieval-Augmented Chain-of-Thought Query Generation}
\begin{figure*}[th!]
    \centering  
    \includegraphics[width=1.0\textwidth]{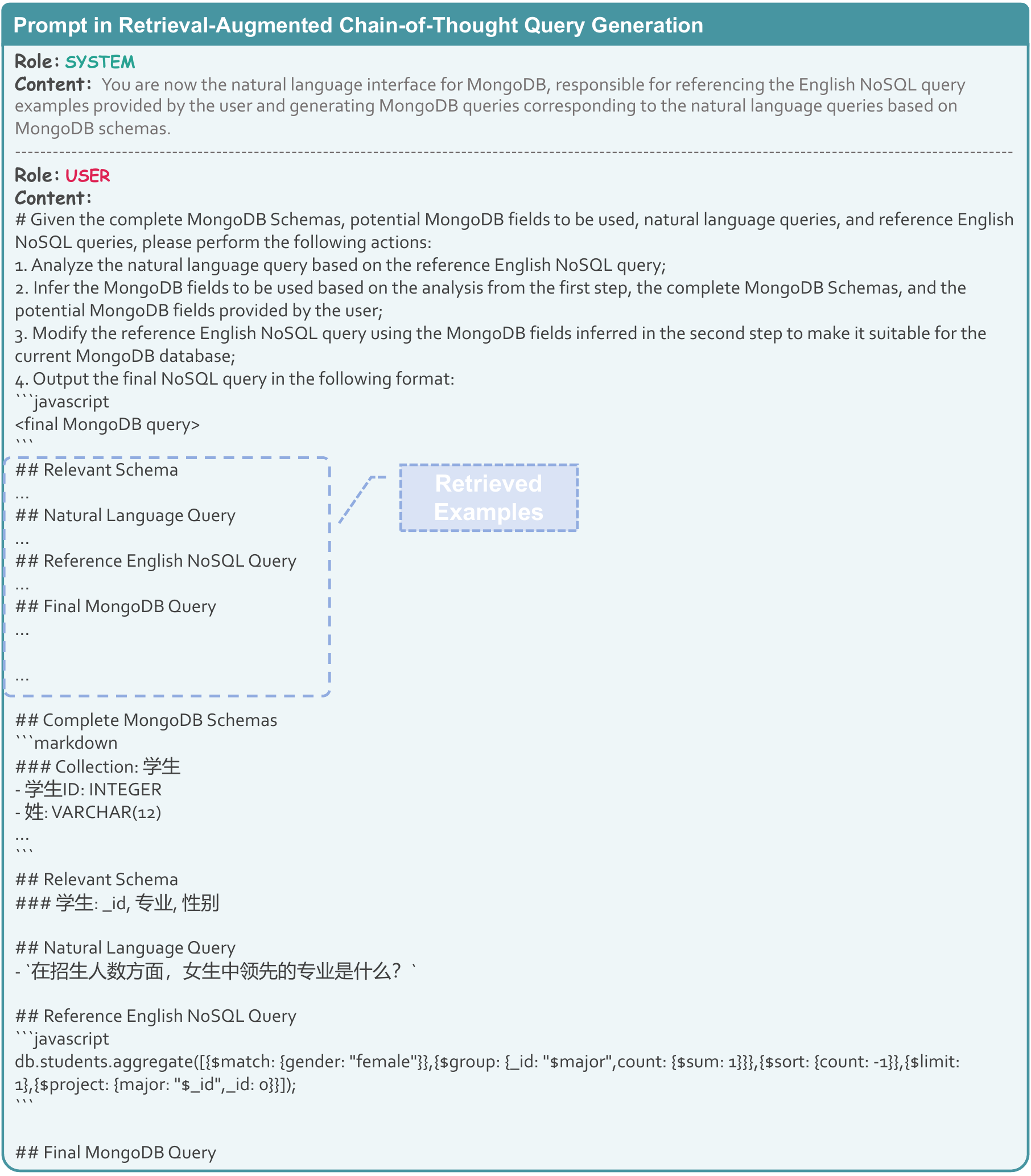}
\end{figure*}

\end{document}